\documentclass[conference]{IEEEtran}
\usepackage{times}
\usepackage{enumitem}

\usepackage[numbers,compress]{natbib}
\usepackage{multicol}
\usepackage[bookmarks=true]{hyperref}
\hypersetup{
	colorlinks=true,
	linkcolor=orange,
	filecolor=magenta,      
	urlcolor=orange,
	citecolor=magenta,
}
\usepackage{xcolor}
\usepackage{amsmath}
\usepackage{graphicx}
\usepackage{booktabs}
\usepackage{multirow}

\usepackage{caption}
\usepackage{subcaption}
\usepackage{etoolbox}

\usepackage[capitalise]{cleveref}
\Crefname{section}{Sec.}{Secs.}
\Crefname{appendix}{App.}{Apps.}

\newcommand{\name}[1]{\textbf{\textit{#1}}}

\definecolor{burntorange}{rgb}{0.8, 0.33, 0.0}
\definecolor{bluegray}{rgb}{0.4, 0.6, 0.8}
\definecolor{asparagus}{rgb}{0.53, 0.66, 0.42}

\usepackage{inconsolata} 
\newcommand{\robotcmd}[1]{%
    \texttt{"#1"}%
}

\begin{document}

\title{Steerable Vision-Language-Action Policies\\for Embodied Reasoning and Hierarchical Control}


\author{
    \authorblockN{
        William Chen\authorrefmark{1},
        Jagdeep Singh Bhatia\authorrefmark{1},
        Catherine Glossop\authorrefmark{1},
        Nikhil Mathihalli\authorrefmark{1},
        Ria Doshi\authorrefmark{2},
        Andy Tang\authorrefmark{2},
        \\
        Danny Driess\authorrefmark{3}, 
        Karl Pertsch\authorrefmark{3},
        Sergey Levine\authorrefmark{1}
    }
    \authorblockA{\authorrefmark{1}U.C. Berkeley, \authorrefmark{2}Stanford University, \authorrefmark{3}Physical Intelligence} 
}

\makeatletter
\let\@oldmaketitle\@maketitle%
\renewcommand{\@maketitle}{\@oldmaketitle%
    \centering
    \includegraphics[width=1\linewidth]{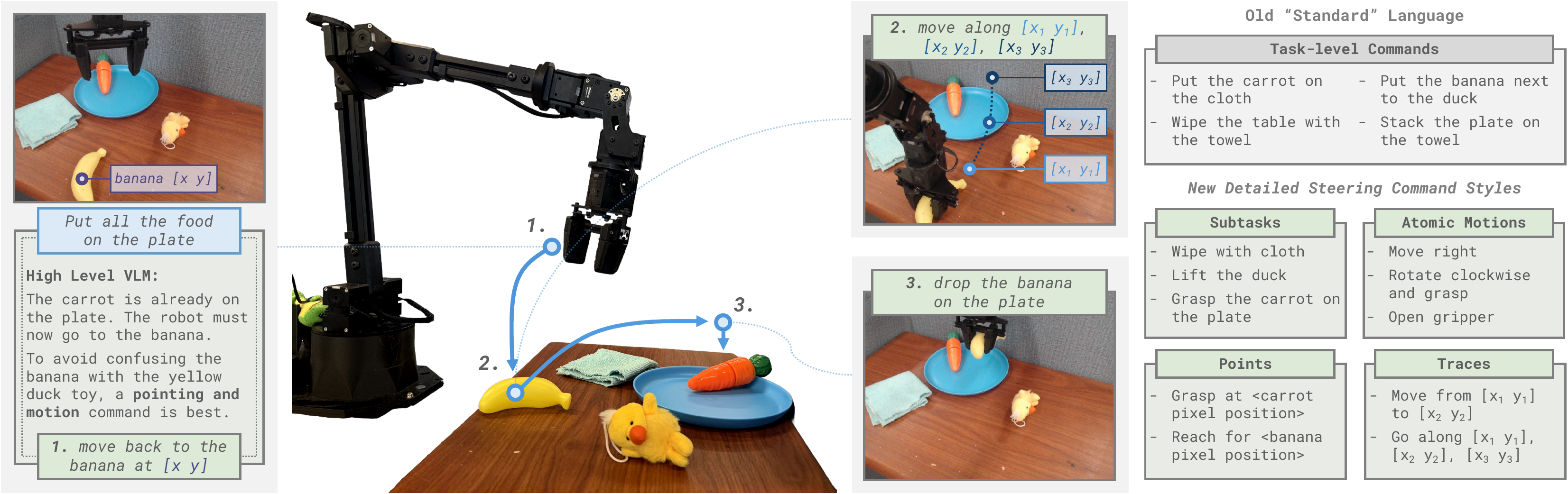}
    \captionof{figure}{We propose \textbf{Steerable Policies}: vision-language-action models that can robustly follow a wide range of detailed commands (green boxes on right), going beyond usual task language to include instructions such as motions or pixel coordinates of the gripper and objects. The flexibility afforded by Steerable Policies enables substantially improved transfer of VLMs' pretrained reasoning, semantic knowledge, and in-context learning skills to generalist robotic control in hierarchical systems.}
    \vspace{-0.5cm}
    \label{fig:teaser}
}

\makeatother
\maketitle
\addtocounter{figure}{-1}
\thispagestyle{empty}
\pagestyle{empty}

\begin{abstract}

    Pretrained vision-language models (VLMs) can make semantic and visual inferences across diverse settings, providing valuable common-sense priors for robotic control. However, effectively grounding this knowledge in robot behaviors remains an open challenge. Prior methods often employ a hierarchical approach where VLMs reason over high-level commands to be executed by separate low-level policies, e.g., vision-language-action models (VLAs). The interface between VLMs and VLAs is usually \emph{natural language task instructions}, which fundamentally limits how much VLM reasoning can steer low-level behavior. We thus introduce Steerable Policies: VLAs trained on rich synthetic commands at various levels of abstraction, like subtasks, motions, and grounded pixel coordinates. By improving low-level controllability, Steerable Policies can unlock pretrained knowledge in VLMs, enabling improved task generalization. We demonstrate this benefit by controlling our Steerable Policies with both a learned high-level embodied reasoner and an off-the-shelf VLM prompted to reason over command abstractions via in-context learning. Across extensive real-world manipulation experiments, these two novel methods outperform prior embodied reasoning VLAs and VLM-based hierarchical baselines, including on challenging generalization and long-horizon tasks.

    \noindent Website: \href{steerable-policies.github.io}{steerable-policies.github.io}   
\end{abstract}

\IEEEpeerreviewmaketitle


\section{Introduction}

A long-standing goal of robotics is to develop \textit{generalist} policies that can follow open-ended commands to perform a wide range of tasks. As scaling robot data collection remains expensive, this motivates the use of pretrained foundation vision-language models (VLMs) as a source of semantic and perceptual knowledge. Prior works often employ a hierarchy, where a high-level VLM reasons over task instructions and issues commands to a separate, language-conditioned low-level policy~\citep{Ahn22-sayCan, Shi25-hiRobot}. However, effectively bringing VLM capabilities to bear for solving general robotics tasks is challenging, as doing so requires grounding their pretrained knowledge in robot behaviors.
This raises a fundamental question: \textit{how can we best unlock and leverage the generalization capabilities of pretrained foundation models for robotics?}

We posit that a major bottleneck in transferring these capabilities is insufficient policy \textit{steerability}. Put simply, no matter how good a VLM is at determining robot behaviors needed for solving tasks, its reasoning abilities are wasted if the robot's policy cannot execute them. For example, the VLM might infer that an object needs to be grasped in a particular location, but this inference is only useful if the policy can understand commands specifying that location. Even powerful policies, such as vision-language-action models (VLA), have limited ability to follow diverse commands due to dataset limitations, where labels are often too formulaic, homogeneous, and imprecise to specify the full range of behaviors needed for solving new tasks~\citep{Wanna26-limitedLinguisticDiversityVLAs}.
These drawbacks make standard task-level language derived from dataset labels a poor interface for VLMs to control robot policies.

One insight is that steerability can be improved by augmenting existing datasets with more detailed synthetic language. Robot datasets already \textit{implicitly} contain rich semantics, interactions, and behaviors, far beyond what is described in their sparse task labels. For example, a policy which learns to ``wipe the table with the towel on the left'' tacitly learns what towels look like, how to grasp them, and how to move left -- all of which can be leveraged for other tasks, like ``hang the towel on the hook.'' 
Training on more descriptive commands could aid the policy flexibly invoke these skills for solving new tasks.

However, na\"ively densifying task descriptions is insufficient for generalization and composition.
For example, expanding the descriptions of existing behaviors would struggle to teach the policy the names of fundamentally out-of-distribution objects.
Still, the policy may be able to grasp these novel objects, due to their physical similarity to objects the policy \textit{has} learned to interact with.
A generalizable way to induce these behaviors is by prompting with \textit{grounded} features, like pixel coordinates, as they can specify behaviors and concepts that are otherwise difficult to communicate to the policy (e.g., \robotcmd{grasp at <novel object's position>}). Not only are grounded features easily extracted for training, but modern VLMs are also trained to output them, thereby allowing these VLMs to transfer their pretrained knowledge to robotics.

We thus introduce \textbf{Steerable Policies}: robotic foundation models that follow a much wider range of commands than standard VLAs. Beyond typical ``task-level'' commands (e.g., \robotcmd{put the carrot in the pot}), we train VLAs to accept more detailed inputs, such as subtasks (\robotcmd{reach for the carrot}) or low-level atomic motions (\robotcmd{move left and close the gripper}). We also include commands with grounded features, such as gripper traces (\robotcmd{move along $[x_1, y_1], [x_2, y_2], ...$}) or points (\robotcmd{grasp the object at $[x, y]$}). Finally, we include prompts that compose all these modalities (\robotcmd{move right from $[x_1, y_1]$ to put the mushroom on the plate at $[x_2, y_2]$}).
These \name{steering commands} are synthetically generated by a scalable pipeline that automatically parses and labels robot data. See \cref{fig:teaser} for more.

Compared to past hierarchical methods~\citep{Ahn22-sayCan}, Steerable Policies vastly expand the interface between low-level robot behaviors and high-level VLM skills, allowing high-level reasoners to flexibly choose instructions at the right level of abstraction for generalization and compositionality. 
We show this by proposing and evaluating two ways for VLMs to control our Steerable Policies, instantiated on a real-world Bridge WidowX setup~\citep{Ebert21-bridgeDataV1, Walke23-bridgeDataV2}.
First, we show that VLMs can be fine-tuned into effective \textit{high-level} models for instructing Steerable Policies by producing chain-of-thought reasonings (CoT~\citep{Wei23-chainOfThought, Kojima23-zeroShotChainOfThought}) followed by steering commands. 
This outperforms past reasoning VLAs~\citep{Zawalski24-ecot, Chen25-ecot-lite}, indicating that it makes especially good use of embodied reasoning data. 
Second, we show how Steerable Policies can better apply the \textit{in-context learning} skills of API-based off-the-shelf VLMs. We find that these VLMs can apply in-context learning over past observations and commands to further improve robot behaviors by choosing commands of the correct level of abstraction.
This is a novel functionality afforded by our Steerable Policies, which leads to significant performance gains over standard baselines.

In summary, we show that improving VLA steerability is critical to facilitate transfer of VLM capabilities to robotics. We do this by introducing \textbf{(1)} Steerable Policies: VLAs that can be prompted at many levels of abstraction to perform diverse manipulation skills -- a paradigm which is agnostic to VLA architecture, as we demonstrate by instantiating Steerable Policies with two popular VLA frameworks (OpenVLA and $\pi_{0.5}$~\citep{Kim24-openVLA, Pi25-pi05}); and \textbf{(2)} novel methods for using VLMs' embodied reasoning and in-context learning to hierarchically control our VLA to solve challenging tasks. We demonstrate that the steerability of low-level VLAs significantly improves the use of high-level VLMs' pretrained capabilities, resulting in better robot generalization.

\section{Related Works}
\label{sec:related-works}

\textbf{Vision-language-action models.} As in vision and language, robotics has moved towards using large foundation models~\citep{Bommasani22-foundationModels}. A popular method is to fine-tune general VLMs into \textit{vision-language-action policies (VLAs)}~\citep{Brohan23-rt2, Kim24-openVLA}, transferring the base VLM's comprehensive pretrained representations to learn robot tasks from demonstrations. Prior works explore many variations of this recipe, such as altering action representations~\citep{Brohan23-rt2, Kim24-openVLA, Pertsch25-fast, Belkhale24-miniVLA, Black24-pi0, Pi25-pi05, Kim25-openVLA-oft} or training VLAs to perform embodied reasoning~\citep{Zawalski24-ecot, Chen25-ecot-lite, GRT25-geminirobotics}. Our approach differs from such works in that we aim to improve policy steerability via augmentation of language prompts, with the end goal of making better use of foundation model capabilities in hierarchical approaches.

\textbf{Training for steerability.} Past works often improve policy steerability by training on synthetic language labels that are more diverse~\citep{Glossop25-cast}, detailed~\citep{Xiao22-robotSkillAcquisitionVLM, Smith24-steer}, or composable~\citep{Zhang24-sprint, Lynch22-interactiveLanguage}. They can also be trained on non-text steering modalities, like gripper traces~\citep{Li25-hamsterhierarchicalactionmodels, Gu23-rttrajectory} or visual marks~\cite{Zheng25-tracevla}. 

In contrast, our method trains on \textit{many} input types, rather than finding a one-size-fits-all steering modality. 
Additionally, all our steering commands are expressed in text tokens, allowing our VLA to interface with generative VLMs more easily than other models trained on many prompt modalities, such as OmniVLA~\citep{Hirose25-omnivla}.
Similar to our method, MolmoAct trains on standard text with gripper trace CoTs, enabling inference-time steering via two modalities (changing either the trace \textit{or} the language)~\citep{Lee25-molmoAct}.
However, unlike our work, they limit language steering to task-level prompts only (no other abstractions, like subtasks, motions, or points, as we use). 
Additionally, MolmoAct's gripper trace steering can only be used \textit{in conjunction} with text commands, which can be a drawback when text introduces adverse biases (e.g., if the policy systematically misidentifies an object by name). This disadvantage is not shared by our policy.

In summary, the core novelty of our Steerable Policies is that they accept steering commands spanning a spectrum of abstractions, not just one or two, as MolmoAct and other prior works do. In particular, giving high-level VLMs the flexibility to \textit{choose} how to steer the low-level VLAs is necessary for improved generalization. We find in \cref{subsec:oracle-commands} that this is critical for performance, as the effectiveness of different steering abstractions depends heavily on the task setting.

\textbf{Synthetic data for model controllability.} Our method of training policies on synthetic language parallels a trend in text-to-image generation: while image corpora often contain the diversity needed for detailed image generation, the accompanying captions are too imprecise for fine-grained text controllability, necessitating detailed synthetic labels~\citep{Betker23-dalle3}. Similarly, our core insight is that \textit{robot datasets already contain diverse behaviors, but their text labels are too sparse, biased, and succinct to induce desirable actions for novel tasks}. 
While steerability in the image domain is valuable for aligning with user preferences, we aim to make VLAs more steerable to better interface with high-level policies, enabling improved use of their pretrained capabilities for solving new control tasks.

\textbf{Hierarchical robot learning.} 
One way to improve generalization is to hierarchically decompose robot tasks into atomic skills~\citep{Garrett20-integratedTAMP}. These techniques broadly divide action prediction into (1) fast and reactive control and (2) slow and methodical reasoning or planning, akin to the System I and II model of human cognition~\citep{Kahneman11-thinkingFastAndSlow}. 
In robot learning, these methods either are trained end-to-end (e.g., policies with an intermediate ``chain-of-thought reasoning''~\citep{Zawalski24-ecot, Pi25-pi05, Lee25-molmoAct, Hwang24-emma}) or, as we focus on in \cref{sec:high-level-methods}, are factorized into separate models~\citep{Ahn22-sayCan, Shi24-yayRobot, Shi25-hiRobot, Li25-hamsterhierarchicalactionmodels}.

Several works use off-the-shelf VLMs as the high-level policy~\citep{Ahn22-sayCan, li2025interactivetaskplanninglanguage, shah2022lmnavroboticnavigationlarge, huang2022languagemodelszeroshotplanners, Huang22-innerMonologue, Zeng22-socraticModels, Driess23-palme, zhang2023bootstrapskillslearningsolve, Ha23-scalingUpDistillingDown}, while others finetune them to improve their alignment with a given domain or low-level policy~\citep{Shi24-yayRobot, Belkhale24-rth, yang2025lohovlaunifiedvisionlanguageactionmodel, peschl2025codeactionhierarchicallearning, Li25-hamsterhierarchicalactionmodels, Lee25-molmoAct, fang2025robixunifiedmodelrobot, jiang2025galaxeaopenworlddatasetg0}. However, these works usually select a \textit{single} modality to interface between high and low levels, such as subtask-level language prompts~\citep{Ahn22-sayCan, huang2022languagemodelszeroshotplanners, Huang22-innerMonologue, peschl2025codeactionhierarchicallearning, fang2025robixunifiedmodelrobot, Driess23-palme, shah2022lmnavroboticnavigationlarge, Zeng22-socraticModels, yang2025lohovlaunifiedvisionlanguageactionmodel, jiang2025galaxeaopenworlddatasetg0} or grounded representations~\citep{Li25-hamsterhierarchicalactionmodels, Lee25-molmoAct}.

Steerable Policies enable a fundamentally new capability absent from these works: the ability to reason about and choose which abstraction to use for steering VLAs. This not only yields high performance (\cref{subsec:oracle-commands}), but also provides novel ways to apply broad VLM capabilities -- scene understanding, reasoning, and in-context learning -- to robotics (\cref{subsec:results-api-vlms}).
Of course, non-steerable VLAs only accept a single type of (usually text) prompt, and do not reap the same benefits.

\begin{figure}[t]
    \centering
    \includegraphics[width=\linewidth]{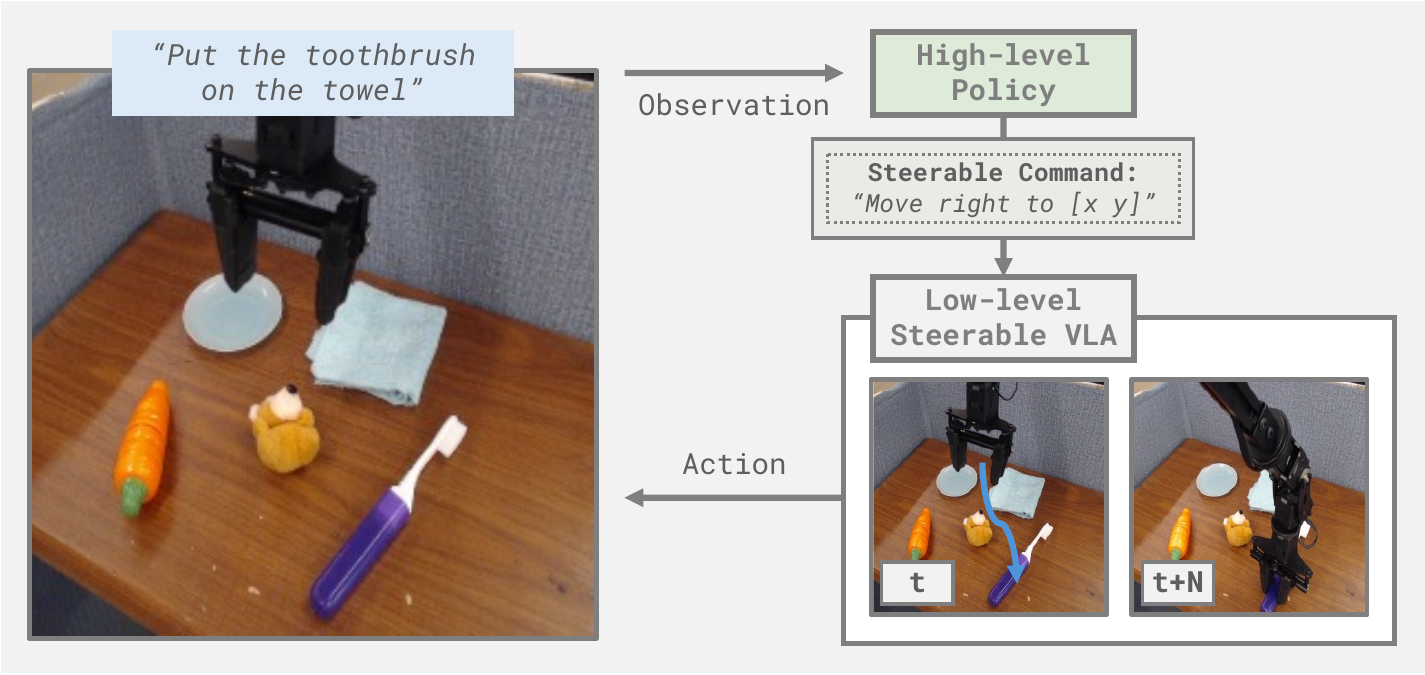}
    \vspace{-0.6cm}
    \caption{The hierarchical policy inference loop, where a high-level model sends commands to the low-level Steerable Policy.}
    \label{fig:hierarchical-loop}
    \vspace{-0.7cm}
\end{figure}

\vspace{-0.1cm}
\section{Preliminaries}
\vspace{-0.1cm}

\textbf{Vision-language-action models.} Given task prompt $l$ and observations $o$ (e.g., proprioception and images), \textit{behavioral cloning (BC)} learns a policy $\pi(a | o, l)$ that solves tasks by matching the behavior distribution of expert demonstrators. \textit{Vision-language-action models} are pretrained \textit{vision-language models} fine-tuned into $\pi$. Our Steerable Policy (\cref{sec:steerable-vlas}) is a VLA, though trained to accept diverse steering modalities.

\textbf{Hierarchical policies.} One way to improve the generalization of learned policies is with hierarchical methods. This factorizes robot instruction-following into two steps: first, a high-level policy takes the overall task $l$ and observation $o$ to output an appropriate intermediate subtask, plan, or goal $g$; then, the low-level policy samples actions $a$ conditioned on $g$ and $o$. This is the framework we adopt (\cref{fig:hierarchical-loop}): our Steerable Policy acts as the low-level controller (where $g$ are steering commands), while we explore the novel high-level policy capabilities they enable in \cref{sec:high-level-methods}.

\section{Steerable Vision-Language-Action Policies}
\label{sec:steerable-vlas}

We now introduce Steerable Policies. We propose training VLAs on \name{steering commands} that span many abstractions, such as subtasks, motions, or grounded coordinates (\cref{subsec:types-of-steerable-commands}).
After generating them at scale (\cref{subsec:generating-steerable-commands}), these prompts randomly replace the standard task-level labels used for BC. Training on steering commands yields a policy that can execute diverse compositional skills when prompted via a range of modalities (\cref{subsec:oracle-commands}), in contrast to past works that use single steering inputs~\citep{Li25-hamsterhierarchicalactionmodels, Gu23-rttrajectory, Zhang24-sprint}. This section solely focuses on training \textit{low-level} Steerable Policies; then, in \cref{sec:high-level-methods}, we discuss how to use these VLAs with \textit{high-level} VLM policies that perform embodied reasoning or in-context learning to choose appropriate steering commands.

\subsection{Styles of Steering Commands}
\label{subsec:types-of-steerable-commands}

To obtain VLAs that can exhibit diverse behavior when instructed with a wide range of abstractions, we train on a diverse mix of different \name{steering command styles}, satisfying several desiderata: commands should be (1) \textit{versatile} for inducing a range of generalizable behaviors for solving novel tasks; (2) \textit{conducive} to being issued by human operators, high-level reasoners, and in-context learning VLMs; and (3) \textit{scalable} for synthetic labeling -- i.e., they can be automatically generated from robot trajectories by using foundation models. 
The categories of steering commands that we train on are as follows (see \cref{fig:teaser} and \cref{app-sec:generating-for-bridge} for more examples):

\begin{enumerate}[leftmargin=*]
    \item \name{Tasks}: The default labels used for training VLAs. They allow Steerable Policies to also solve tasks that standard VLAs can. E.g., \robotcmd{put the carrot in the pot}.
    \item \name{Subtasks}: These help compose existing semantic skills for novel tasks. E.g., \robotcmd{reach for the carrot}.
    \item \name{Atomic motions}: These let the VLA follow granular movements in language, without referencing the semantic contents of the scene. E.g., \robotcmd{move left} or \robotcmd{open gripper}.
    \item \name{Gripper traces}: These provide a list of pixels for the gripper to follow. E.g., \robotcmd{move from $[x_1, y_1]$ to $[x_2, y_2]$}.
    \item \name{Points}: These indicate positions of objects or locations of interest. E.g., \robotcmd{lift above <pot position>} or \robotcmd{grasp at <carrot position>}. This is subtly distinct from gripper trace commands, as pointing does \textit{not} necessarily indicate the exact pixels the gripper must move to, merely visual positions relevant to the task.
    \item \name{Combinations}: Hybrids of these styles. E.g., \robotcmd{move left from $[x, y]$ to the carrot at <carrot position>}.
\end{enumerate}

The last three styles can reference 2D pixel coordinates, which VLMs can easily specify~\citep{Deitke24-molmo-pixmo, GRT25-geminirobotics}. They also allow fine-grained steering when language is insufficient: e.g., the policy might fail to pick up some out-of-distribution object specified by name, but would succeed if told to \robotcmd{pick up the object at $[x, y]$}. Likewise, if there are multiple of instances an object (making language underspecified), the policy can determine the correct one to interact with from a pointing command. 

\begin{figure}[t]
    \centering
    \includegraphics[width=1\linewidth]{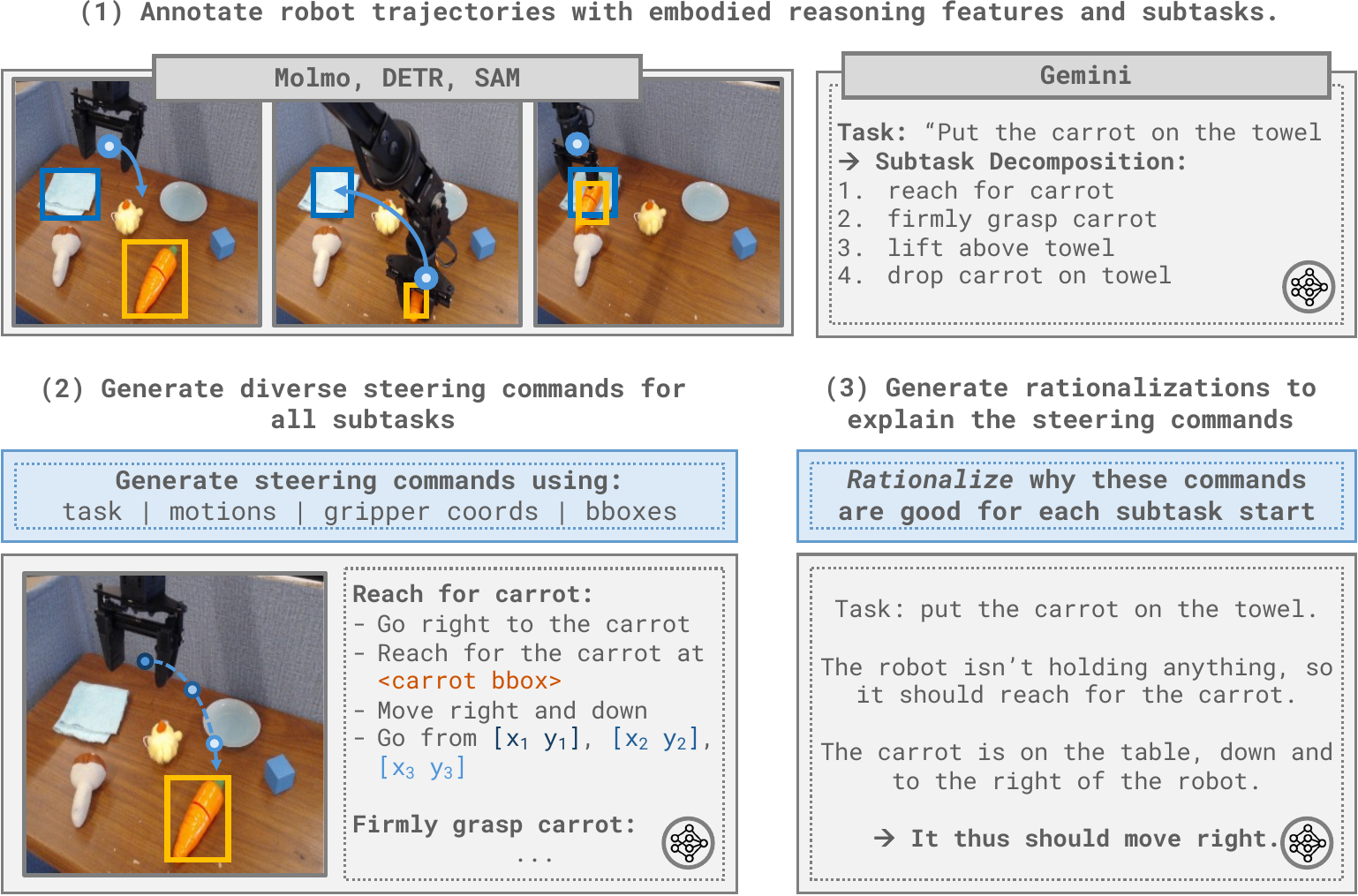}
    \vspace{-0.4cm}
    \caption{Our automated pipeline for annotating robot data with synthetic steering commands at scale. 
    \textbf{1:} We use a suite of foundation models to extract subtasks and grounded features (bounding boxes, motions, and gripper traces) from each trajectory.
    \textbf{2:} We query a VLM to generate diverse steering commands for training Steerable Policies. These commands may reference features extracted in the first step, which we provide in the prompt.
    \textbf{3:} To train high-level embodied reasoners, we also generate \textit{rationalizations} for why particular commands are appropriate for given observations (\cref{subsec:method-learned-embodied-reasoners}).}
    \label{fig:command-pipeline}
    \vspace{-0.7cm}
\end{figure}

\vspace{-0.1cm}
\subsection{Generating Steering Commands at Scale}
\label{subsec:generating-steerable-commands}

The default task commands found in large robot datasets are usually human-annotated. While this yields the gold standard in terms of data quality, acquiring these labels at scale is expensive and labor-intensive, especially when language commands reference grounded features, such as pixel coordinates. We thus turn to synthetically-generated annotations.

Our pipeline for producing steering commands is shown in \cref{fig:command-pipeline}. We first extract grounded features from a robot dataset, then compile them into subtasks.
Specifically, we follow \citet{Zawalski24-ecot} to programmatically extract motion language, then use Molmo to extract task-relevant object names and map them to segmentation masks~\citep{Deitke24-molmo-pixmo}. Next, we use SAM2 to track and propagate these masks across the entire trajectory video~\citep{Ravi24-sam2}, thereby producing temporally-consistent open-vocabulary bounding boxes. Separately, we call DETR to extract the robot's gripper traces~\citep{Carion20-detr}. Finally, given the overall task, motions, and objects, we query Gemini 2.0~\citep{GeminiTeam24-gemini} to break the episode into semantic subtasks.

After grounded feature extraction and subtask decomposition, we query Gemini \textit{again} to restate each subtask into equivalent steering commands of all styles in \cref{subsec:types-of-steerable-commands}). Gripper traces, motions, and object centroids are provided in the prompt, as these grounded features are useful for generating some steering commands.
This pipeline allows us to expand from 38k ``standard'' Bridge task-level language labels to 206k subtasks and nearly 2M total steering commands. See \cref{app-sec:generating-for-bridge} for full details and all prompts used.

The generated commands are used to train Steerable Policies by simply replacing the standard language during BC with a corresponding steering command (sampled uniformly at random across all commands for each frame). The resulting VLA can thus follow all command styles in \cref{subsec:types-of-steerable-commands} necessary to produce generalizable behaviors, which we confirm in \cref{subsec:oracle-commands}. In particular, we show that different command styles are suited to different tasks and situations, suggesting Steerable Polices generalize better. Indeed, we also find that choosing steering commands intelligently yields a substantial performance boost over a non-steerable baseline.

\vspace{-0.05cm}
\section{Hierarchical Methods with Steerable Policies}
\label{sec:high-level-methods}
\vspace{-0.05cm}

\begin{figure}[t]
    \centering
    \includegraphics[width=\linewidth]{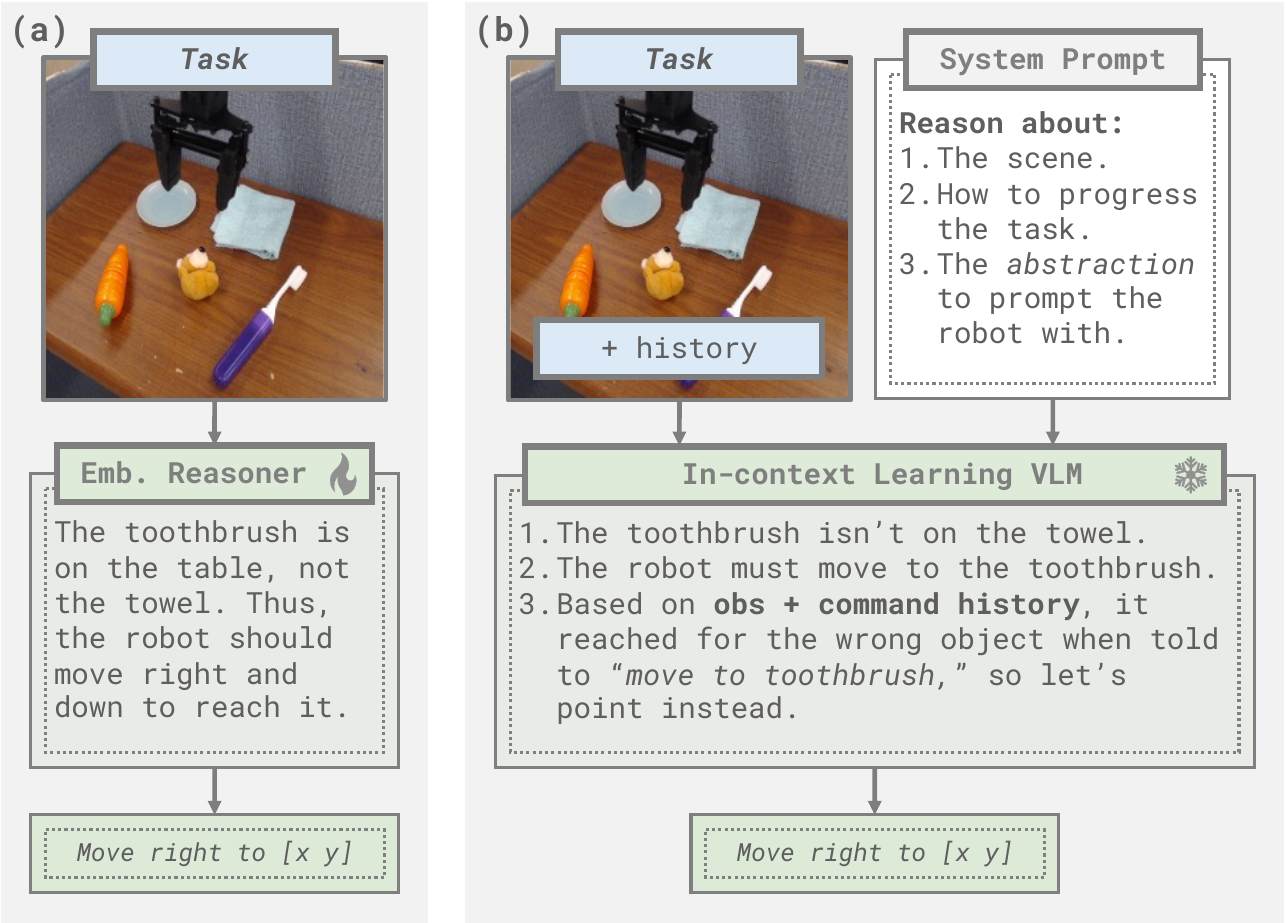}
    \vspace{-0.6cm}
    \caption{Our two novel high-level policies (\cref{sec:high-level-methods}). \textbf{(a)} fine-tunes a VLM into an embodied reasoner that issues steering commands, while \textbf{(b)} queries an off-the-shelf VLM to determine appropriate commands via in-context reasoning.}
    \label{fig:hierarchical_methods}
    \vspace{-0.7cm}
\end{figure}

To fully realize the benefits of Steerable Policies, they must be integrated with VLMs that understand their affordances. We introduce two \textit{high-level} VLM policies that effectively leverage semantic knowledge and physical reasoning to issue steering commands. 
First, we investigate how embodied reasoning training can aid lightweight open-source VLMs to leverage their pretrained representations for controlling Steerable Policies. Second, we use the zero-shot in-context learning abilities of off-the-shelf VLMs for multi-step robotic problem-solving. Both methods are depicted in \cref{fig:hierarchical_methods}.

\vspace{-0.05cm}
\subsection{Training High-level Embodied Reasoners}
\label{subsec:method-learned-embodied-reasoners}
\vspace{-0.05cm}

Our first method trains a high-level embodied reasoning model to decompose tasks into steering commands, as shown in \cref{fig:hierarchical_methods} (a). Specifically, we fine-tune a VLM to output appropriate steering commands for accomplishing a given task based on a task instruction and the current observation. 
Note that our command generation pipeline (\cref{subsec:generating-steerable-commands}) provides a mapping from tasks to appropriate commands, yielding the exact supervision necessary to train this model.

To make better use of the base VLMs' reasoning abilities and improve generalization, we also produce post-hoc rationales for \textit{why} each command is appropriate. For each subtask in every trajectory, we query Gemini with its starting frame, asking the VLM to explain why that subtask is needed to make progress (\cref{fig:command-pipeline}). See \cref{app-subsec:generating-rationalizatons} for the full prompt used and more details. We then train the high-level policy via next-token prediction to reason about what the robot should do in each frame. During inference, the model autoregressively predicts a reasoning followed by a corresponding steering command, which our Steerable Policy follows for $N=5$ environment steps before re-querying the high-level reasoner.

We expect this approach to be particularly effective for three main reasons. First, as the high-level reasoner's learning objective (mapping from task language to natural language reasonings and steering commands) is similar to the base VLM's pretraining objective, we expect it to readily transfer its generalizable representations to this new task. Second, since the low-level VLA only attends to steering commands, it learns fewer spurious biases between the task, reasoning, and action, compared to most unfactorized end-to-end models.
Finally, since the high-level reasoner is queried less frequently than the VLA, it speeds up inference compared to standard embodied reasoning, even without any compilation techniques~\citep{Chen25-tensorrt-openvla, Nvidia-tensorRT-LLM}.

\subsection{In-context Reasoning for Choosing Steering Abstractions}
\label{subsec:method-api-vlms}

Steerable Policies also enable a novel functionality: they permit many command styles, letting high-level models \textit{choose} which level of abstraction is best for inducing behaviors that progress a given task. 
Thus, in addition to reasoning about \textit{what} the robot should do, hierarchical systems using Steerable Policies have the added dimension of reasoning about \textit{how} to get the robot to do it. 
Our approach in \cref{subsec:method-learned-embodied-reasoners} fine-tunes a model to map tasks to steering commands, but not necessarily those of the optimal style. Doing so requires (1) intuition about the strengths and weaknesses of each command style and (2) the ability to learn when each style is effective on the fly. We thus hypothesize that \textit{off-the-shelf} VLMs would excel at selecting command styles, due to their zero-shot proficiency with in-context learning and reasoning.

We test this idea with the implementation shown in \cref{fig:hierarchical_methods} (b). We query an API-based off-the-shelf VLM with the robot's observation and an overall task. The VLM is told to issue commands to our Steerable Policy, based on examples of all the command styles and brief descriptions of their strengths and weaknesses. 
The model receives a history of images and executed commands in context, and is asked to (1) parse the current scene, (2) determine what the robot should do next, and (3) reason about the best level of abstraction for commanding the policy.
Finally, it predicts a steering command for the Steerable Policy to follow for $N=20$ steps before the high-level VLM is re-queried. See \cref{app-sec:api-vlms-details} for more details.

Beyond emitting steering commands at a range of abstraction styles, our approach has several key innovations. Most significantly, the VLM can use \textit{in-context learning} to adapt its command choices, since it receives a sequential history of past observations and commands it has chosen. After selecting a steering abstraction, the VLM can observe the robot's resulting behavior, then adjust the abstraction or command as needed. 
Critically, because in-context data points are produced by the VLM, it does not need hand-crafted examples of robot behaviors.
Note that as in-context learning is used here to predict steering commands, not robot actions, the VLM's task is thus akin to the ``standard'' vision-language in-context learning that VLMs are known to excel at. This obviates domain-specialized training or structured scene descriptions needed in prior robotic in-context learning works~\citep{Fu24-icrt, Yin25-robotInContextLearningLLMs}. 

Another benefit of our approach is that the VLM can use its scene understanding in conjunction with descriptions of each command style to reason about optimal steering abstractions. For instance, the VLM might produce visual inferences such as ``the scene is cluttered,'' leading to the selection of a pointing or motion command which allows for improved specificity. 

Naturally, as the above benefits stem from choosing steering abstractions, they are \textit{not} applicable to prior methods where VLMs control standard policies that accept only a single conditioning modality (e.g., task language~\citep{Brohan23-rt2, Kim24-openVLA, Zhang24-sprint} or traces~\citep{Li25-hamsterhierarchicalactionmodels, Gu23-rttrajectory, Zheng25-tracevla}). Only Steerable Policies can fully bring these VLM capabilities to bear for robotic problem-solving.
\section{Experiments}
\label{sec:experiments}

\begin{figure}[t]
    \centering
    \vspace{0.2cm}
    \includegraphics[width=1\linewidth]{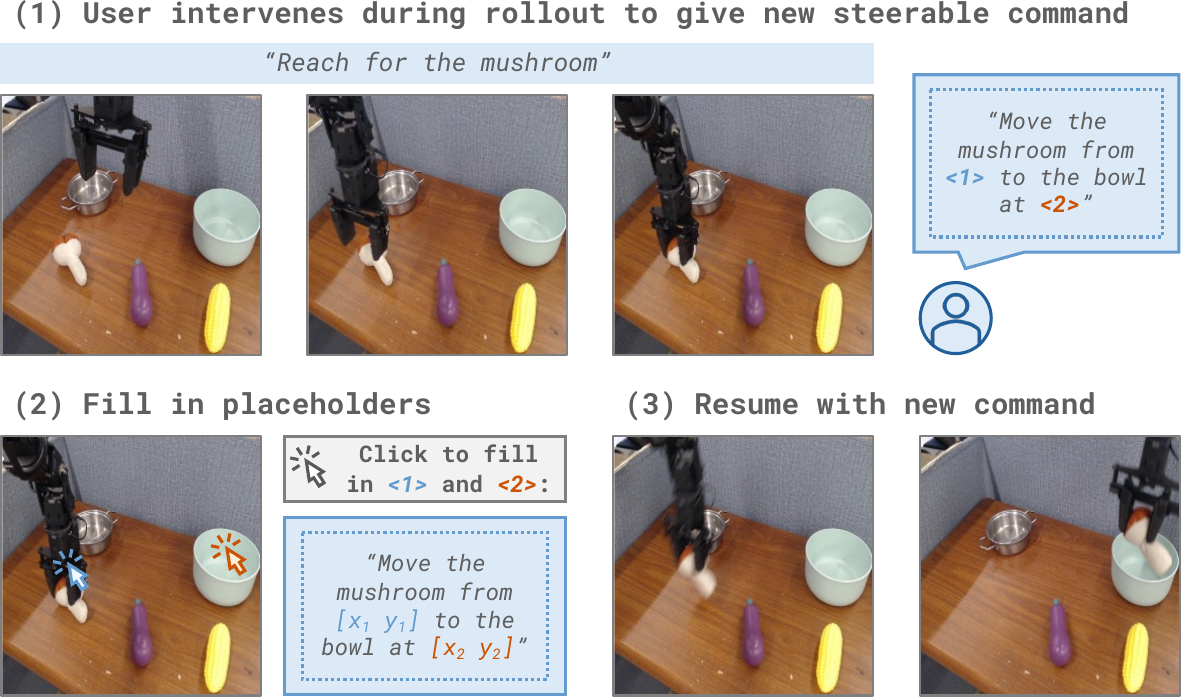}
    \vspace{-0.6cm}
    \caption{Interactive interface for querying humans for oracle steering commands. \textbf{1:} The operator can interrupt the rollout to issue a new steering command. To facilitate giving commands with pixel coordinates, they can add textual placeholder markers. \textbf{2:} If any are given, a GUI is opened displaying the current robot observation, allowing the user to click to fill the markers in. \textbf{3:} Finally, the rollout resumes with the new command.}
    \label{fig:intervention-diagram}

    \centering
    \includegraphics[width=1\linewidth]{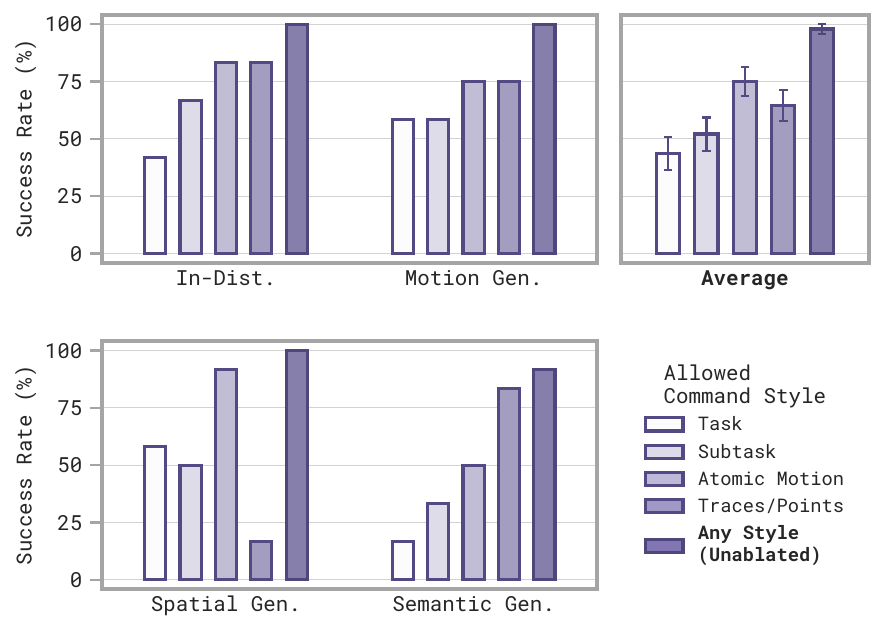}
    \vspace{-0.7cm}
    \caption{Allowing the oracle human user to issue \textit{any} command style to our Steerable Policy \textbf{nearly saturates performance} on Bridge. By restricting the user to each style alone, we find each one is suited to different task types. All individual styles are better than directly providing the task-level label that is used by regular VLAs. Error bars denote $\pm 1$StdErr.}
    \label{fig:command-type-results}
    \vspace{-0.8cm}
\end{figure}

\begin{figure*}[t]
    \centering
    \includegraphics[width=1\linewidth]{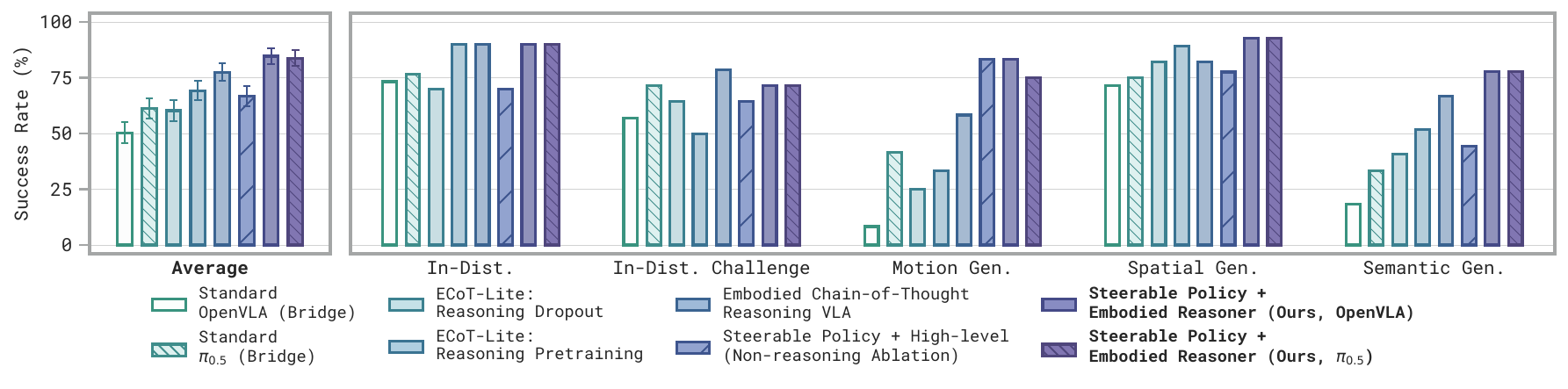}
    \vspace{-0.8cm}
    \caption{
    Our approach of controlling Steerable Policies with learned high-level embodied reasoning VLMs outperforms five baselines: the equivalent standard Bridge OpenVLA and $\pi_{0.5}$~\citep{Kim24-openVLA, Pi25-pi05}, the Reasoning Pretraining and Dropout ECoT-Lite methods~\citep{Chen25-ecot-lite}, and full Embodied Chain-of-Thought Reasoning~\citep{Zawalski24-ecot}. Error bars denote $\pm 1$StdErr. We adopt the same Bridge task suite as ECoT-Lite~\citep{Chen25-ecot-lite}, enabling a direct comparison with other methods for training VLAs with embodied reasoning data.}
    \label{fig:reasoner-results}

    \includegraphics[width=1\linewidth]{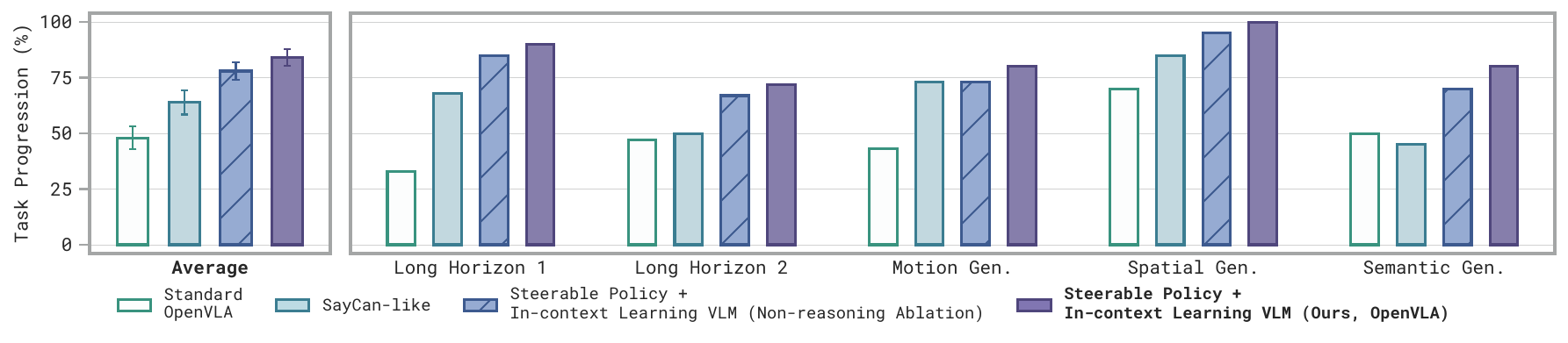}
    \vspace{-0.8cm}
    \caption{In-context learning VLMs can effectively select abstractions for instructing our Steerable Policy. Error bars denote $\pm 1$StdErr. Steerability allows VLMs to better apply their reasoning and in-context learning skills, in ways the standard SayCan-like paradigm (where the VLM interfaces with the VLA via subtasks alone) fails to take advantage of (\cref{subsec:results-api-vlms}). To better highlight these capabilities, we measure task progression on challenging \textit{multi-step} Bridge tasks.}
    \label{fig:multiabstraction-results}
    \vspace{-0.7cm}
\end{figure*}

\textbf{Research questions.} We now evaluate our Steerable Policies, both individually and as part of our two hierarchical control methods.
We aim to answer: 
\textbf{(1)} Can steering commands effectively induce diverse compositional behaviors in Steerable Policies, yielding improved task performance?
\textbf{(2)} How can Steerable Policies enable trained high-level VLMs to leverage embodied reasoning data for generalization?
\textbf{(3)} How can Steerable Policies let us better apply in-context learning, scene understanding, and reasoning competencies in off-the-shelf VLMs for solving long-horizon tasks?

\textbf{Experimental setup.} We train Steerable Policies on the Bridge WidowX real-world robot manipulation dataset~\citep{Ebert21-bridgeDataV1, Walke23-bridgeDataV2, EmbodimentCollaboration24-oxe}, allowing for evaluations consistent with past generalist policies~\citep{OMT23-octo, Kim24-openVLA, Zawalski24-ecot, Chen25-ecot-lite}. Following these works, we adapt the OpenVLA codebase to train our Steerable Policy (and embodied reasoner) with the standard Prismatic VLM 7B architecture~\citep{Karamcheti24-prismatic, Kim24-openVLA}, but modified to map \textit{steering commands} (rather than standard task labels) and third-person RGB images to end effector actions. To show that Steerable Policies are applicable to many VLAs, we also fine-tune one from $\pi_{0.5}$~\citep{Pi25-pi05}, which we use in \cref{subsec:results-learned-embodied-reasoners}. See \cref{app-sec:training-details} for more details. 

For experiments in \cref{subsec:oracle-commands} and \cref{subsec:results-learned-embodied-reasoners}, we measure performance over common generalization axes~\citep{Kim24-openVLA, Zawalski24-ecot, Chen25-ecot-lite}: in-distribution, motion, spatial, and semantic generalization. When evaluating our in-context learning high-level policy formulation in \cref{subsec:results-api-vlms}, we additionally include unseen longer-horizon tasks. See \cref{app-sec:experiment-details} for full task details.

\subsection{Diverse Steering Commands Induce Performant and\\Compositional Behaviors in Steerable Policies}
\label{subsec:oracle-commands}

To improve the performance of learned hierarchical systems, the underlying policies must respond to a wider range of commands beyond standard task-level language.
Our first experiment evaluates whether our proposed steering commands are sufficiently expressive to induce the compositional behaviors necessary for Steerable Policies to solve a range of novel tasks.

Specifically, we establish an achievable upper bound on performance by having a human act as an ``oracle'' high-level policy. This experiment serves as a didactic proof of existence showing that, given human-level scene understanding and reasoning, steering commands enable a Steerable Policy to attain high task performance. Following \citet{Shi24-yayRobot}, the user may intervene by modifying the policy's free-form steering command as needed (\cref{fig:intervention-diagram}). However, they must wait $\geq 2$ seconds between interventions to reflect practical hierarchical systems that query the high-level policy less frequently than the low-level. This also prevents the user from ``fully teleoperating'' the robot by repeatedly changing the command.

We additionally repeat this experiment while restricting the user to a single command style at a time, including the `task-level'' prompts used by standard VLAs. This isolates whether \textit{individual} steering modalities are sufficient for strong performance, and clarifies \textit{when} each style is most effective.

Our results demonstrate that \textbf{a human oracle issuing unrestricted steering commands can solve all tasks at nearly 100\% success rate} (\cref{fig:command-type-results}). Even when constrained to a single command style, the user consistently outperforms task-level language alone. Additionally, we find that no single style dominates across settings: each exhibits complementary strengths and weaknesses. Optimal performance empirically requires access to a spectrum of prompt abstraction, supporting the claim that no one-size-fits-all steering modality exists.

We observe several trends for when each style is effective. First, \name{trace/point} commands dominate in semantic generalization, as pointing allows the policy to know which out-of-distribution objects to interact with, even if it fails to identify them by name. Next, \name{atomic motion} commands are best in tasks that reference spatial relations. When following motion commands, the policy remains on a manifold of ``reasonable'' actions -- e.g., when prompted to ``move left,'' it servos left \textit{towards an object}, rather than moving left unconditionally (see \cref{app-subsec:manifold-of-reasonable-actions} for more examples). As a result, when using a single steering modality, motions are the most effective. Finally, while \name{task/subtask} commands never perform best in any evaluation, they are especially reliable for \textit{in-distribution} parts of an overall task (e.g., reaching for non-novel objects).

Overall, we find our Steerable Policy to be both responsive to different command styles and highly performant, particularly when given suitable steering commands leveraging \textit{all} command styles. However, selecting effective commands requires reasoning about the task, the scene, and the VLA's capabilities. This motivates the novel hierarchical methods presented in \cref{sec:high-level-methods}, which we evaluate below.

\subsection{Steerable Policies Enable VLMs to Effectively Use\\Embodied Reasoning Training}
\label{subsec:results-learned-embodied-reasoners}

We now evaluate if fine-tuning a VLM into a high-level policy is an effective way to leverage embodied reasoning to control Steerable Policies. 
Since our method involves fine-tuning the high-level to produce embodied reasonings and intermediate goals, we compare against three prior methods that use reasoning data to improve VLAs (all built on OpenVLA): \name{Embodied Chain-of-Thought Reasoning (ECoT)}~\citep{Zawalski24-ecot} and the two \name{ECoT-Lite} variants, which learn representations from robot reasonings, but do not generate them during inference~\citep{Chen25-ecot-lite}. We also include a non-reasoning ablation, where the fine-tuned VLM outputs steering commands directly. Finally, we compare against equivalent non-reasoning \name{``standard'' OpenVLA and $\pi_{0.5}$ baselines}~\citep{Kim24-openVLA, Pi25-pi05}. All methods are evaluated on the same Bridge tasks used in ECoT-Lite, which were chosen to probe several axes of generalization. For fairness, we control for training demonstrations and architectures; the policies differ only in how embodied reasoning is incorporated.

As shown in \cref{fig:reasoner-results}, \textit{our approach outperforms all baselines}. The gains are most pronounced on motion and semantic generalization tasks, likely because atomic motion, trace, and pointing commands allow the generalizable reasoner to reliably control the Steerable Policy under distribution shift.

\textbf{High-level VLM reasoning benefits Steerable Policies based on different VLA architectures.} While the ``standard'' Bridge $\pi_{0.5}$ baseline outperforms the OpenVLA equivalent (likely due to having better base representations for control), applying our method to $\pi_{0.5}$- \textit{or} OpenVLA-based Steerable Policies still outperforms both baselines, showing that \textit{steerability unlocks performance gains from VLM reasoning, even for modern VLA architectures}. Both Steerable Policies achieve similar performance when using our approach.

\textbf{Fine-tuned high-level VLMs benefit from reasoning.} While not as effective as our full approach, ablating reasoning from our method still greatly outperforms standard OpenVLA and is on-par with ECoT-Lite, despite being trained on the same robot interactions. This suggests that there are performance benefits to hierarchical control schemes with Steerable Policies over end-to-end policies, \textit{even if the high-level is not explicitly trained to reason}. Still, we find that embodied reasoning is important for generalization and performance. Note that VLMs' off-the-shelf reasoning abilities can be used to achieve the same benefit, which we discuss below.

\subsection{Steerable Policies Unlock In-context Reasoning in VLMs}
\label{subsec:results-api-vlms}

\begin{figure}[h]
    \begingroup

    \newcommand{\SubCapAbove}{2pt}
    \newcommand{\SubCapBelow}{3pt}
    \newcommand{\FigCapAbove}{0pt}
    \newcommand{\FigCapBelow}{0pt}

    \captionsetup[subfigure]{
        aboveskip=\SubCapAbove,
        belowskip=\SubCapBelow
    }
    \captionsetup{
        aboveskip=\FigCapAbove,
        belowskip=\FigCapBelow
    }

    \centering

    \begin{subfigure}[b]{\linewidth}
         \centering
         \includegraphics[width=\linewidth]{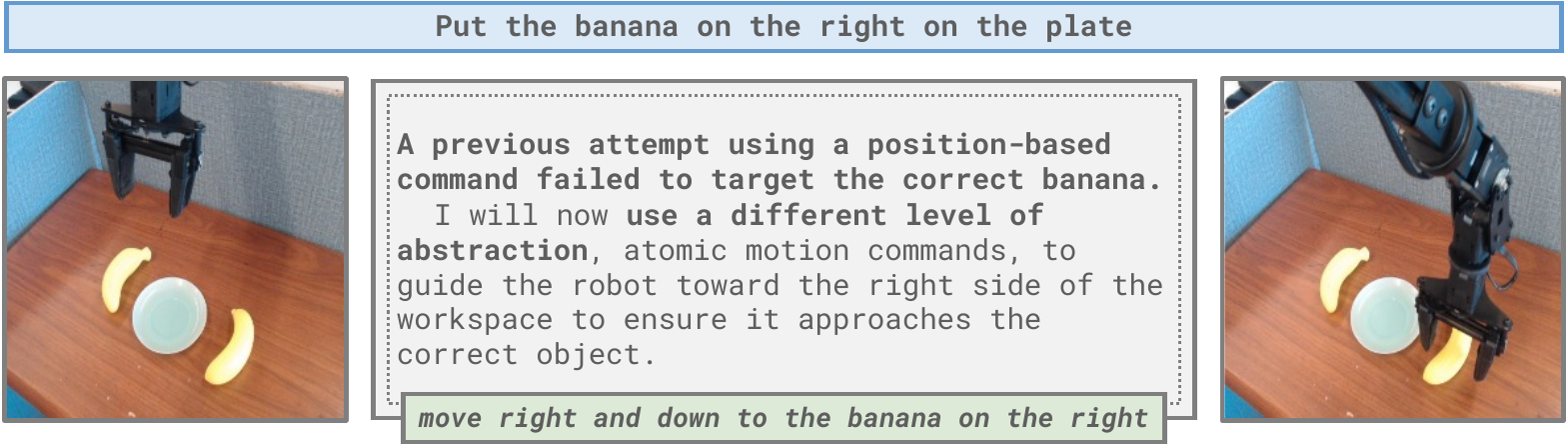}
         \caption{The VLM uses in-context learning to ``discover'' what steering abstraction is best. A pointing command leads to incorrect behavior, so the VLM employs a motion command instead.}
         \label{fig:vlm-examples-3}
     \end{subfigure}

     \begin{subfigure}[b]{\linewidth}
         \centering
         \includegraphics[width=\linewidth]{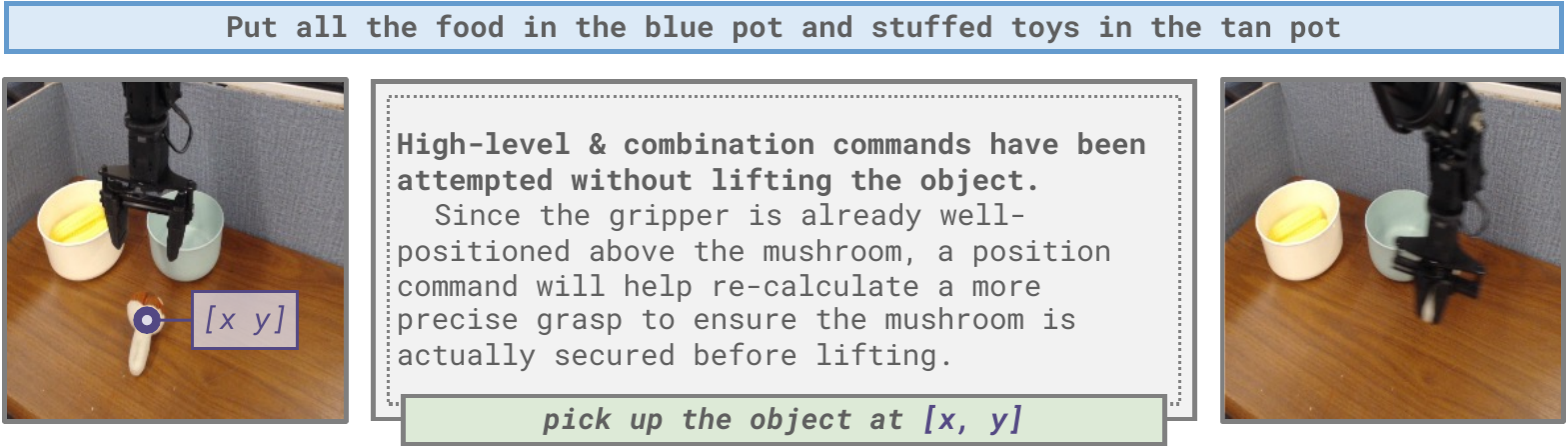}
         \caption{The VLM uses in-context learning to adjust the abstraction when task progress stalls. Subtask (\robotcmd{lift the mushroom up}) and motion (\robotcmd{move down and close gripper onto the mushroom}) commands were ineffective, so the VLM points instead.}
         \label{fig:vlm-examples-4}
     \end{subfigure}

    \begin{subfigure}[b]{\linewidth}
         \centering
         \includegraphics[width=\linewidth]{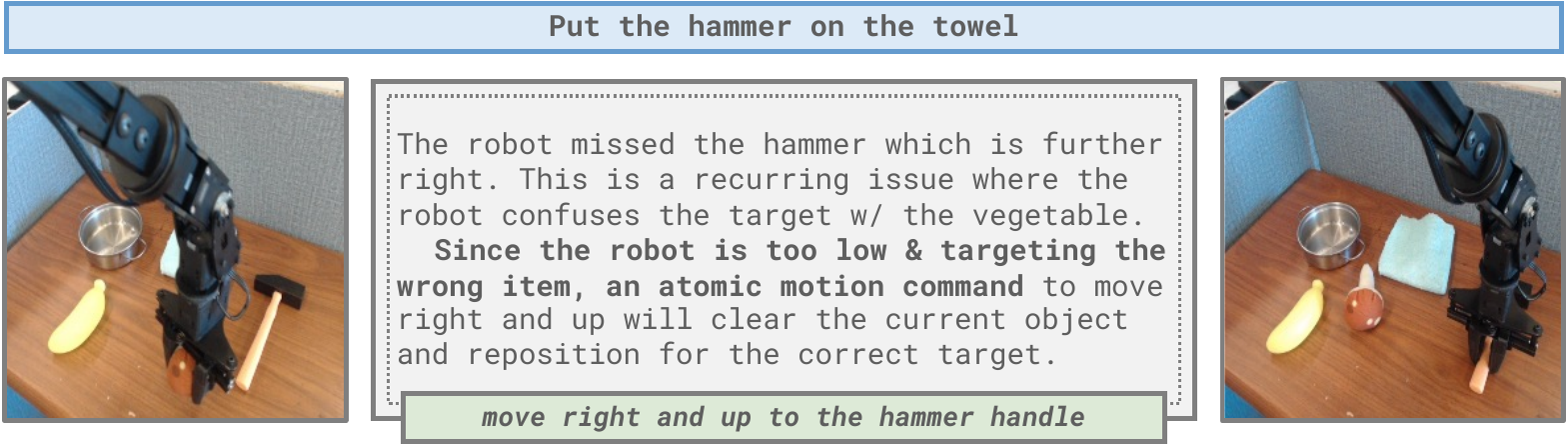}
         \caption{The VLM demonstrates fine-grained semantic and physical understanding to issue a corrective command.}
         \label{fig:vlm-examples-1}
     \end{subfigure}

     \begin{subfigure}[b]{\linewidth}
         \centering
         \includegraphics[width=\linewidth]{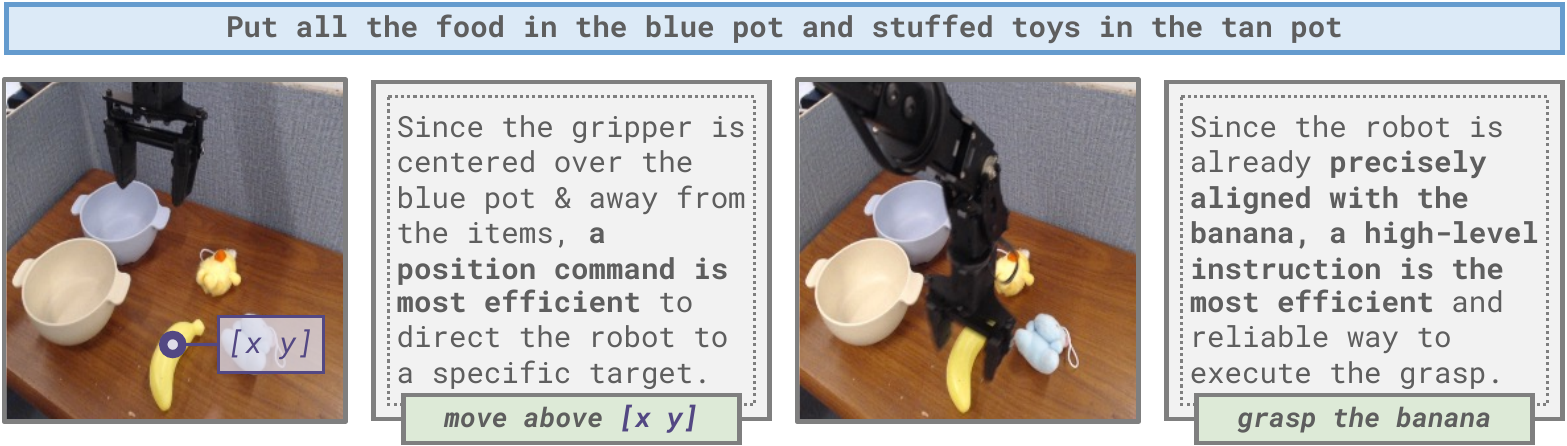}
         \caption{Given an observation, the VLM reasons about which steering abstraction is most appropriate, based on each one's strengths.}
         \label{fig:vlm-examples-2}
     \end{subfigure}

    \caption{Example reasonings when using an in-context learning VLM for controlling our Steerable Policies.}
    \label{fig:vlm-examples}
    \vspace{-0.8cm}
    \endgroup
\end{figure}

Finally, we investigate if off-the-shelf VLMs can better use their in-context reasoning faculties when interfacing with Steerable Policies. We define a distinct suite of multi-step tasks that demand reasoning over long-horizon executions. See \cref{app-subsec:experiments-off-the-shelf} for task details and rubrics.

The key advantage of using Steerable Policies is that they allow high-level VLMs to reason about and flexibly choose the best abstraction for prompting low-level VLAs. Our primary baseline is therefore a \name{SayCan-like}~\citep{Ahn22-sayCan} method, representing a standard approach where an off-the-shelf VLM commands the VLA through subtask language alone. 
For fairness, this baseline uses the same history and system prompt as our method, differing only in that the VLM is restricted to a single level of steering abstraction. We expect this to hinder the impact of in-context reasoning, lowering performance.
We also include a \name{standard Bridge OpenVLA} comparison to assess if our tasks are sufficiently challenging for non-hierarchical policies. Finally, to isolate the role of explicit reasoning, we include a \name{non-reasoning ablation} where the VLM outputs steering commands without any intermediate generation. All hierarchical methods use Gemini 3.0~\citep{GeminiTeam24-gemini} as the VLM. More details and prompts for all methods are shown in \cref{app-sec:api-vlms-details}.

As shown in \cref{fig:multiabstraction-results}, our approach universally outperforms OpenVLA and SayCan-like baselines. Since the latter differs only in being restricted to subtask-level prompts, the performance gap primarily reflects the benefit of allowing the VLM to access the full spectrum of steering abstractions during in-context reasoning. Our method also outperforms the non-reasoning ablation, though the gap is smaller. This suggests that, while explicit reasoning is helpful, in-context learning VLMs are already adept at issuing effective steering commands without intermediate ``thinking.'' We further observe that using in-context reasoning to steer our VLA leads to several qualitative advantages, shown below and in \cref{fig:vlm-examples}.

\textbf{In-context learning enables corrective steering.} Even when the VLM initially picks poor steering commands, we find it can use in-context learning to improve commands by changing the abstraction level. This is especially salient when a command leads to an error (\cref{fig:vlm-examples-3}) or is ineffective and fails to progress the task (\cref{fig:vlm-examples-4}). In either case, the VLM reasons over history to learn the affordances of the VLA and change its prompting strategy to complete the task.

\textbf{Steerability better applies VLM understanding.}
Interfacing with a Steerable Policy allows the VLM to translate its scene understanding and in-context history into actionable corrections that are difficult to induce in a standard VLA accepting task-level language alone. In \cref{fig:vlm-examples-1}, the VLM uses \textit{semantic} understanding to observe that the robot grasps the wrong object when instructed to reach for the hammer. It further reasons with granular \textit{physical} understanding, recognizing that the end effector must have sufficient clearance to disengage the current object and move above the correct one. It uses this understanding in conjunction with past behavioral and semantic failure modes to ground its final steering command: \robotcmd{move right and up to the hammer handle}.

Crucially, the VLM often cannot capitalize on such insights with subtask commands alone. A common failure mode of the SayCan-like baseline is that, even when the VLM correctly diagnoses errors, subtasks lack the specificity required to resolve them. In the same example (\cref{fig:vlm-examples-1}), the VLM may recognize that the robot has confused a mushroom for the hammer and issue \robotcmd{drop the mushroom}. However, since a non-steerable VLA cannot understand motion commands such as \robotcmd{move up and right}, the VLM can only reissue the subtask command \robotcmd{reach for the hammer}, causing the robot to repeat the same mistake.
In short, even if the VLM has the capacity for nuanced scene understanding, it cannot act on this knowledge with subtask commands alone, requiring a low-level Steerable Policy instead (see \cref{app-subsec:subtasks-insufficient}).

\textbf{VLMs can reason about steering abstractions.} Beyond reasoning about \textit{what} the robot should do, VLMs can also reason about \textit{how} to best induce the desired behavior (i.e., what steering style to use). In \cref{fig:vlm-examples-2}, when reaching for a banana in a cluttered scene, the VLM determines that a pointing command is the least ambiguous option and issues \robotcmd{move above <banana position>}. After observing that the robot is correctly above the banana, it switches to a higher-level instruction, \robotcmd{grasp the banana}. This demonstrates that VLMs can dynamically select and sequence steering abstractions based on the current state.

\section{Discussion and Future Work}
We introduce Steerable Policies: VLAs that support a wide range of command abstractions including subtasks, motions, and grounded pixel coordinates. We show how improved steerability enables better transfer of VLM capabilities into robotics by (1) fine-tuning high-level VLMs on embodied reasoning data to solve generalization tasks with our Steerable Policy, and (2) using the in-context learning abilities of off-the-shelf VLMs for multi-step robotic problem-solving.


As robot datasets have trended toward scaling their behavioral diversity, we expect Steerable Policies to only increase in applicability.
The compositionality afforded by steering also becomes more important as task complexity increases, e.g., when moving to open-world settings. To support this, VLMs require an even better grasp of the affordances provided by Steerable Policies. Beyond zero-shot prompting, VLMs could \textit{learn} these affordances via reinforcement learning, as rollouts would allow models to elucidate when each steering style is effective. This may also provide the data needed for cross-task in-context learning, where the VLM transfers its understanding of affordances to new situations. We hope that Steerable Policies will represent a step toward finally bringing the ``disembodied'' capabilities of VLMs to the physical world.

\section*{Acknowledgements}
\noindent We thank Qiyang Li for assistance with the teaser diagram. 
We also thank Ameesh Shah, Arhan Jain, and members of the RAIL Lab for insightful discussions.
This research was partly supported by ONR N00014-25-1-2060, NSF IIS-2246811, and DARPA ANSR, with additional support from Google and the NVIDIA Academic Grant Program.


\bibliographystyle{unsrtnat}
\bibliography{references}

@misc{Ahn22-sayCan,
      title={Do As I Can, Not As I Say: Grounding Language in Robotic Affordances}, 
      author={Michael Ahn and Anthony Brohan and Noah Brown and Yevgen Chebotar and Omar Cortes and Byron David and Chelsea Finn and Chuyuan Fu and Keerthana Gopalakrishnan and Karol Hausman and Alex Herzog and Daniel Ho and Jasmine Hsu and Julian Ibarz and Brian Ichter and Alex Irpan and Eric Jang and Rosario Jauregui Ruano and Kyle Jeffrey and Sally Jesmonth and Nikhil J Joshi and Ryan Julian and Dmitry Kalashnikov and Yuheng Kuang and Kuang-Huei Lee and Sergey Levine and Yao Lu and Linda Luu and Carolina Parada and Peter Pastor and Jornell Quiambao and Kanishka Rao and Jarek Rettinghouse and Diego Reyes and Pierre Sermanet and Nicolas Sievers and Clayton Tan and Alexander Toshev and Vincent Vanhoucke and Fei Xia and Ted Xiao and Peng Xu and Sichun Xu and Mengyuan Yan and Andy Zeng},
      year={2022},
      eprint={2204.01691},
      archivePrefix={arXiv},
      primaryClass={cs.RO}
}

@misc{Brohan23-rt2,
      title={RT-2: Vision-Language-Action Models Transfer Web Knowledge to Robotic Control}, 
      author={Anthony Brohan and Noah Brown and Justice Carbajal and Yevgen Chebotar and Xi Chen and Krzysztof Choromanski and Tianli Ding and Danny Driess and Avinava Dubey and Chelsea Finn and Pete Florence and Chuyuan Fu and Montse Gonzalez Arenas and Keerthana Gopalakrishnan and Kehang Han and Karol Hausman and Alexander Herzog and Jasmine Hsu and Brian Ichter and Alex Irpan and Nikhil Joshi and Ryan Julian and Dmitry Kalashnikov and Yuheng Kuang and Isabel Leal and Lisa Lee and Tsang-Wei Edward Lee and Sergey Levine and Yao Lu and Henryk Michalewski and Igor Mordatch and Karl Pertsch and Kanishka Rao and Krista Reymann and Michael Ryoo and Grecia Salazar and Pannag Sanketi and Pierre Sermanet and Jaspiar Singh and Anikait Singh and Radu Soricut and Huong Tran and Vincent Vanhoucke and Quan Vuong and Ayzaan Wahid and Stefan Welker and Paul Wohlhart and Jialin Wu and Fei Xia and Ted Xiao and Peng Xu and Sichun Xu and Tianhe Yu and Brianna Zitkovich},
      year={2023},
      eprint={2307.15818},
      archivePrefix={arXiv},
      primaryClass={cs.RO}
}

@misc{Driess23-palme,
      title={PaLM-E: An Embodied Multimodal Language Model}, 
      author={Danny Driess and Fei Xia and Mehdi S. M. Sajjadi and Corey Lynch and Aakanksha Chowdhery and Brian Ichter and Ayzaan Wahid and Jonathan Tompson and Quan Vuong and Tianhe Yu and Wenlong Huang and Yevgen Chebotar and Pierre Sermanet and Daniel Duckworth and Sergey Levine and Vincent Vanhoucke and Karol Hausman and Marc Toussaint and Klaus Greff and Andy Zeng and Igor Mordatch and Pete Florence},
      year={2023},
      eprint={2303.03378},
      archivePrefix={arXiv},
      primaryClass={cs.LG}
}

@misc{Huang22-innerMonologue,
      title={Inner Monologue: Embodied Reasoning through Planning with Language Models}, 
      author={Wenlong Huang and Fei Xia and Ted Xiao and Harris Chan and Jacky Liang and Pete Florence and Andy Zeng and Jonathan Tompson and Igor Mordatch and Yevgen Chebotar and Pierre Sermanet and Noah Brown and Tomas Jackson and Linda Luu and Sergey Levine and Karol Hausman and Brian Ichter},
      year={2022},
      eprint={2207.05608},
      archivePrefix={arXiv},
      primaryClass={cs.RO}
}

@misc{Karamcheti24-prismatic,
      title={Prismatic VLMs: Investigating the Design Space of Visually-Conditioned Language Models}, 
      author={Siddharth Karamcheti and Suraj Nair and Ashwin Balakrishna and Percy Liang and Thomas Kollar and Dorsa Sadigh},
      year={2024},
      eprint={2402.07865},
      archivePrefix={arXiv},
      primaryClass={cs.CV}
}

@misc{Garrett20-integratedTAMP,
      title={Integrated Task and Motion Planning}, 
      author={Caelan Reed Garrett and Rohan Chitnis and Rachel Holladay and Beomjoon Kim and Tom Silver and Leslie Pack Kaelbling and Tomás Lozano-Pérez},
      year={2020},
      eprint={2010.01083},
      archivePrefix={arXiv},
      primaryClass={cs.RO}
}

@misc{Wei23-chainOfThought,
      title={Chain-of-Thought Prompting Elicits Reasoning in Large Language Models}, 
      author={Jason Wei and Xuezhi Wang and Dale Schuurmans and Maarten Bosma and Brian Ichter and Fei Xia and Ed Chi and Quoc Le and Denny Zhou},
      year={2023},
      eprint={2201.11903},
      archivePrefix={arXiv},
      primaryClass={cs.CL}
}

@misc{Kojima23-zeroShotChainOfThought,
      title={Large Language Models are Zero-Shot Reasoners}, 
      author={Takeshi Kojima and Shixiang Shane Gu and Machel Reid and Yutaka Matsuo and Yusuke Iwasawa},
      year={2023},
      eprint={2205.11916},
      archivePrefix={arXiv},
      primaryClass={cs.CL}
}

@misc{Zeng22-socraticModels,
    title={Socratic Models: Composing Zero-Shot Multimodal Reasoning with Language}, 
    author={Andy Zeng and Maria Attarian and Brian Ichter and Krzysztof Choromanski and Adrian Wong and Stefan Welker and Federico Tombari and Aveek Purohit and Michael Ryoo and Vikas Sindhwani and Johnny Lee and Vincent Vanhoucke and Pete Florence},
    year={2022}}

@misc{Ravi24-sam2,
      title={SAM 2: Segment Anything in Images and Videos}, 
      author={Nikhila Ravi and Valentin Gabeur and Yuan-Ting Hu and Ronghang Hu and Chaitanya Ryali and Tengyu Ma and Haitham Khedr and Roman Rädle and Chloe Rolland and Laura Gustafson and Eric Mintun and Junting Pan and Kalyan Vasudev Alwala and Nicolas Carion and Chao-Yuan Wu and Ross Girshick and Piotr Dollár and Christoph Feichtenhofer},
      year={2024},
      eprint={2408.00714},
      archivePrefix={arXiv},
      primaryClass={cs.CV},
      url={https://arxiv.org/abs/2408.00714}, 
}

@misc{GeminiTeam24-gemini,
      title={Gemini: A Family of Highly Capable Multimodal Models}, 
      author={{Gemini Team}},
      year={2024},
      eprint={2312.11805},
      archivePrefix={arXiv},
      primaryClass={cs.CL}
}

@inproceedings{Walke23-bridgeDataV2,
    title={BridgeData V2: A Dataset for Robot Learning at Scale},
    author={Walke, Homer and Black, Kevin and Lee, Abraham and Kim, Moo Jin and Du, Max and Zheng, Chongyi and Zhao, Tony and Hansen-Estruch, Philippe and Vuong, Quan and He, Andre and Myers, Vivek and Fang, Kuan and Finn, Chelsea and Levine, Sergey},
    booktitle={Conference on Robot Learning (CoRL)},
    year={2023}
}

@misc{EmbodimentCollaboration24-oxe,
      title={Open X-Embodiment: Robotic Learning Datasets and RT-X Models}, 
      author={{Embodiment Collaboration} and Abby O'Neill and Abdul Rehman and Abhiram Maddukuri and Abhishek Gupta and Abhishek Padalkar and Abraham Lee and Acorn Pooley and Agrim Gupta and Ajay Mandlekar and Ajinkya Jain and Albert Tung and Alex Bewley and Alex Herzog and Alex Irpan and Alexander Khazatsky and Anant Rai and Anchit Gupta and Andrew Wang and Andrey Kolobov and Anikait Singh and Animesh Garg and Aniruddha Kembhavi and Annie Xie and Anthony Brohan and Antonin Raffin and Archit Sharma and Arefeh Yavary and Arhan Jain and Ashwin Balakrishna and Ayzaan Wahid and Ben Burgess-Limerick and Beomjoon Kim and Bernhard Schölkopf and Blake Wulfe and Brian Ichter and Cewu Lu and Charles Xu and Charlotte Le and Chelsea Finn and Chen Wang and Chenfeng Xu and Cheng Chi and Chenguang Huang and Christine Chan and Christopher Agia and Chuer Pan and Chuyuan Fu and Coline Devin and Danfei Xu and Daniel Morton and Danny Driess and Daphne Chen and Deepak Pathak and Dhruv Shah and Dieter Büchler and Dinesh Jayaraman and Dmitry Kalashnikov and Dorsa Sadigh and Edward Johns and Ethan Foster and Fangchen Liu and Federico Ceola and Fei Xia and Feiyu Zhao and Felipe Vieira Frujeri and Freek Stulp and Gaoyue Zhou and Gaurav S. Sukhatme and Gautam Salhotra and Ge Yan and Gilbert Feng and Giulio Schiavi and Glen Berseth and Gregory Kahn and Guanzhi Wang and Hao Su and Hao-Shu Fang and Haochen Shi and Henghui Bao and Heni Ben Amor and Henrik I Christensen and Hiroki Furuta and Homer Walke and Hongjie Fang and Huy Ha and Igor Mordatch and Ilija Radosavovic and Isabel Leal and Jacky Liang and Jad Abou-Chakra and Jaehyung Kim and Jaimyn Drake and Jan Peters and Jan Schneider and Jasmine Hsu and Jeannette Bohg and Jeffrey Bingham and Jeffrey Wu and Jensen Gao and Jiaheng Hu and Jiajun Wu and Jialin Wu and Jiankai Sun and Jianlan Luo and Jiayuan Gu and Jie Tan and Jihoon Oh and Jimmy Wu and Jingpei Lu and Jingyun Yang and Jitendra Malik and João Silvério and Joey Hejna and Jonathan Booher and Jonathan Tompson and Jonathan Yang and Jordi Salvador and Joseph J. Lim and Junhyek Han and Kaiyuan Wang and Kanishka Rao and Karl Pertsch and Karol Hausman and Keegan Go and Keerthana Gopalakrishnan and Ken Goldberg and Kendra Byrne and Kenneth Oslund and Kento Kawaharazuka and Kevin Black and Kevin Lin and Kevin Zhang and Kiana Ehsani and Kiran Lekkala and Kirsty Ellis and Krishan Rana and Krishnan Srinivasan and Kuan Fang and Kunal Pratap Singh and Kuo-Hao Zeng and Kyle Hatch and Kyle Hsu and Laurent Itti and Lawrence Yunliang Chen and Lerrel Pinto and Li Fei-Fei and Liam Tan and Linxi "Jim" Fan and Lionel Ott and Lisa Lee and Luca Weihs and Magnum Chen and Marion Lepert and Marius Memmel and Masayoshi Tomizuka and Masha Itkina and Mateo Guaman Castro and Max Spero and Maximilian Du and Michael Ahn and Michael C. Yip and Mingtong Zhang and Mingyu Ding and Minho Heo and Mohan Kumar Srirama and Mohit Sharma and Moo Jin Kim and Naoaki Kanazawa and Nicklas Hansen and Nicolas Heess and Nikhil J Joshi and Niko Suenderhauf and Ning Liu and Norman Di Palo and Nur Muhammad Mahi Shafiullah and Oier Mees and Oliver Kroemer and Osbert Bastani and Pannag R Sanketi and Patrick "Tree" Miller and Patrick Yin and Paul Wohlhart and Peng Xu and Peter David Fagan and Peter Mitrano and Pierre Sermanet and Pieter Abbeel and Priya Sundaresan and Qiuyu Chen and Quan Vuong and Rafael Rafailov and Ran Tian and Ria Doshi and Roberto Mart{'i}n-Mart{'i}n and Rohan Baijal and Rosario Scalise and Rose Hendrix and Roy Lin and Runjia Qian and Ruohan Zhang and Russell Mendonca and Rutav Shah and Ryan Hoque and Ryan Julian and Samuel Bustamante and Sean Kirmani and Sergey Levine and Shan Lin and Sherry Moore and Shikhar Bahl and Shivin Dass and Shubham Sonawani and Shuran Song and Sichun Xu and Siddhant Haldar and Siddharth Karamcheti and Simeon Adebola and Simon Guist and Soroush Nasiriany and Stefan Schaal and Stefan Welker and Stephen Tian and Subramanian Ramamoorthy and Sudeep Dasari and Suneel Belkhale and Sungjae Park and Suraj Nair and Suvir Mirchandani and Takayuki Osa and Tanmay Gupta and Tatsuya Harada and Tatsuya Matsushima and Ted Xiao and Thomas Kollar and Tianhe Yu and Tianli Ding and Todor Davchev and Tony Z. Zhao and Travis Armstrong and Trevor Darrell and Trinity Chung and Vidhi Jain and Vincent Vanhoucke and Wei Zhan and Wenxuan Zhou and Wolfram Burgard and Xi Chen and Xiangyu Chen and Xiaolong Wang and Xinghao Zhu and Xinyang Geng and Xiyuan Liu and Xu Liangwei and Xuanlin Li and Yansong Pang and Yao Lu and Yecheng Jason Ma and Yejin Kim and Yevgen Chebotar and Yifan Zhou and Yifeng Zhu and Yilin Wu and Ying Xu and Yixuan Wang and Yonatan Bisk and Yoonyoung Cho and Youngwoon Lee and Yuchen Cui and Yue Cao and Yueh-Hua Wu and Yujin Tang and Yuke Zhu and Yunchu Zhang and Yunfan Jiang and Yunshuang Li and Yunzhu Li and Yusuke Iwasawa and Yutaka Matsuo and Zehan Ma and Zhuo Xu and Zichen Jeff Cui and Zichen Zhang and Zipeng Fu and Zipeng Lin},
      year={2024},
      eprint={2310.08864},
      archivePrefix={arXiv},
      primaryClass={cs.RO}
}

@inproceedings{Kim24-openVLA,
    title={OpenVLA: An Open-Source Vision-Language-Action Model},
    author={{Moo Jin} Kim and Karl Pertsch and Siddharth Karamcheti and Ted Xiao and Ashwin Balakrishna and Suraj Nair and Rafael Rafailov and Ethan Foster and Pannag Sanketi and Quan Vuong and Thomas Kollar and Benjamin Burchfiel and Russ Tedrake and Dorsa Sadigh and Sergey Levine and Percy Liang and Chelsea Finn},
    journal = {arXiv preprint},
    year={2024},
}

@misc{Zhai23-siglip,
      title={Sigmoid Loss for Language Image Pre-Training}, 
      author={Xiaohua Zhai and Basil Mustafa and Alexander Kolesnikov and Lucas Beyer},
      year={2023},
      eprint={2303.15343},
      archivePrefix={arXiv},
      primaryClass={cs.CV}
}

@inproceedings{OMT23-octo,
    title={Octo: An Open-Source Generalist Robot Policy},
    author = {{Octo Model Team} and Dibya Ghosh and Homer Walke and Karl Pertsch and Kevin Black and Oier Mees and Sudeep Dasari and Joey Hejna and Charles Xu and Jianlan Luo and Tobias Kreiman and {You Liang} Tan and Lawrence Yunliang Chen and Pannag Sanketi and Quan Vuong and Ted Xiao and Dorsa Sadigh and Chelsea Finn and Sergey Levine},
    booktitle = {Proceedings of Robotics: Science and Systems},
    address  = {Delft, Netherlands},
    year = {2024},
}

@misc{Ebert21-bridgeDataV1,
      title={Bridge Data: Boosting Generalization of Robotic Skills with Cross-Domain Datasets}, 
      author={Frederik Ebert and Yanlai Yang and Karl Schmeckpeper and Bernadette Bucher and Georgios Georgakis and Kostas Daniilidis and Chelsea Finn and Sergey Levine},
      year={2021},
      eprint={2109.13396},
      archivePrefix={arXiv},
      primaryClass={cs.RO}
}

@article{Khazatsky24-droid,
    title   = {DROID: A Large-Scale In-The-Wild Robot Manipulation Dataset},
    author  = {Alexander Khazatsky and Karl Pertsch and Suraj Nair and Ashwin Balakrishna and Sudeep Dasari and Siddharth Karamcheti and Soroush Nasiriany and Mohan Kumar Srirama and Lawrence Yunliang Chen and Kirsty Ellis and Peter David Fagan and Joey Hejna and Masha Itkina and Marion Lepert and Yecheng Jason Ma and Patrick Tree Miller and Jimmy Wu and Suneel Belkhale and Shivin Dass and Huy Ha and Arhan Jain and Abraham Lee and Youngwoon Lee and Marius Memmel and Sungjae Park and Ilija Radosavovic and Kaiyuan Wang and Albert Zhan and Kevin Black and Cheng Chi and Kyle Beltran Hatch and Shan Lin and Jingpei Lu and Jean Mercat and Abdul Rehman and Pannag R Sanketi and Archit Sharma and Cody Simpson and Quan Vuong and Homer Rich Walke and Blake Wulfe and Ted Xiao and Jonathan Heewon Yang and Arefeh Yavary and Tony Z. Zhao and Christopher Agia and Rohan Baijal and Mateo Guaman Castro and Daphne Chen and Qiuyu Chen and Trinity Chung and Jaimyn Drake and Ethan Paul Foster and Jensen Gao and David Antonio Herrera and Minho Heo and Kyle Hsu and Jiaheng Hu and Donovon Jackson and Charlotte Le and Yunshuang Li and Kevin Lin and Roy Lin and Zehan Ma and Abhiram Maddukuri and Suvir Mirchandani and Daniel Morton and Tony Nguyen and Abigail O'Neill and Rosario Scalise and Derick Seale and Victor Son and Stephen Tian and Emi Tran and Andrew E. Wang and Yilin Wu and Annie Xie and Jingyun Yang and Patrick Yin and Yunchu Zhang and Osbert Bastani and Glen Berseth and Jeannette Bohg and Ken Goldberg and Abhinav Gupta and Abhishek Gupta and Dinesh Jayaraman and Joseph J Lim and Jitendra Malik and Roberto Martín-Martín and Subramanian Ramamoorthy and Dorsa Sadigh and Shuran Song and Jiajun Wu and Michael C. Yip and Yuke Zhu and Thomas Kollar and Sergey Levine and Chelsea Finn},
    year    = {2024},
}

@misc{Bommasani22-foundationModels,
      title={On the Opportunities and Risks of Foundation Models}, 
      author={Rishi Bommasani and Drew A. Hudson and Ehsan Adeli and Russ Altman and Simran Arora and Sydney von Arx and Michael S. Bernstein and Jeannette Bohg and Antoine Bosselut and Emma Brunskill and Erik Brynjolfsson and Shyamal Buch and Dallas Card and Rodrigo Castellon and Niladri Chatterji and Annie Chen and Kathleen Creel and Jared Quincy Davis and Dora Demszky and Chris Donahue and Moussa Doumbouya and Esin Durmus and Stefano Ermon and John Etchemendy and Kawin Ethayarajh and Li Fei-Fei and Chelsea Finn and Trevor Gale and Lauren Gillespie and Karan Goel and Noah Goodman and Shelby Grossman and Neel Guha and Tatsunori Hashimoto and Peter Henderson and John Hewitt and Daniel E. Ho and Jenny Hong and Kyle Hsu and Jing Huang and Thomas Icard and Saahil Jain and Dan Jurafsky and Pratyusha Kalluri and Siddharth Karamcheti and Geoff Keeling and Fereshte Khani and Omar Khattab and Pang Wei Koh and Mark Krass and Ranjay Krishna and Rohith Kuditipudi and Ananya Kumar and Faisal Ladhak and Mina Lee and Tony Lee and Jure Leskovec and Isabelle Levent and Xiang Lisa Li and Xuechen Li and Tengyu Ma and Ali Malik and Christopher D. Manning and Suvir Mirchandani and Eric Mitchell and Zanele Munyikwa and Suraj Nair and Avanika Narayan and Deepak Narayanan and Ben Newman and Allen Nie and Juan Carlos Niebles and Hamed Nilforoshan and Julian Nyarko and Giray Ogut and Laurel Orr and Isabel Papadimitriou and Joon Sung Park and Chris Piech and Eva Portelance and Christopher Potts and Aditi Raghunathan and Rob Reich and Hongyu Ren and Frieda Rong and Yusuf Roohani and Camilo Ruiz and Jack Ryan and Christopher Ré and Dorsa Sadigh and Shiori Sagawa and Keshav Santhanam and Andy Shih and Krishnan Srinivasan and Alex Tamkin and Rohan Taori and Armin W. Thomas and Florian Tramèr and Rose E. Wang and William Wang and Bohan Wu and Jiajun Wu and Yuhuai Wu and Sang Michael Xie and Michihiro Yasunaga and Jiaxuan You and Matei Zaharia and Michael Zhang and Tianyi Zhang and Xikun Zhang and Yuhui Zhang and Lucia Zheng and Kaitlyn Zhou and Percy Liang},
      year={2022},
      eprint={2108.07258},
      archivePrefix={arXiv},
      primaryClass={cs.LG}
}

@misc{Oquab24-dinov2,
      title={DINOv2: Learning Robust Visual Features without Supervision}, 
      author={Maxime Oquab and Timothée Darcet and Théo Moutakanni and Huy Vo and Marc Szafraniec and Vasil Khalidov and Pierre Fernandez and Daniel Haziza and Francisco Massa and Alaaeldin El-Nouby and Mahmoud Assran and Nicolas Ballas and Wojciech Galuba and Russell Howes and Po-Yao Huang and Shang-Wen Li and Ishan Misra and Michael Rabbat and Vasu Sharma and Gabriel Synnaeve and Hu Xu and Hervé Jegou and Julien Mairal and Patrick Labatut and Armand Joulin and Piotr Bojanowski},
      year={2024},
      eprint={2304.07193},
      archivePrefix={arXiv},
      primaryClass={cs.CV}
}

@misc{Touvron23-llama2,
      title={Llama 2: Open Foundation and Fine-Tuned Chat Models}, 
      author={Hugo Touvron and Louis Martin and Kevin Stone and Peter Albert and Amjad Almahairi and Yasmine Babaei and Nikolay Bashlykov and Soumya Batra and Prajjwal Bhargava and Shruti Bhosale and Dan Bikel and Lukas Blecher and Cristian Canton Ferrer and Moya Chen and Guillem Cucurull and David Esiobu and Jude Fernandes and Jeremy Fu and Wenyin Fu and Brian Fuller and Cynthia Gao and Vedanuj Goswami and Naman Goyal and Anthony Hartshorn and Saghar Hosseini and Rui Hou and Hakan Inan and Marcin Kardas and Viktor Kerkez and Madian Khabsa and Isabel Kloumann and Artem Korenev and Punit Singh Koura and Marie-Anne Lachaux and Thibaut Lavril and Jenya Lee and Diana Liskovich and Yinghai Lu and Yuning Mao and Xavier Martinet and Todor Mihaylov and Pushkar Mishra and Igor Molybog and Yixin Nie and Andrew Poulton and Jeremy Reizenstein and Rashi Rungta and Kalyan Saladi and Alan Schelten and Ruan Silva and Eric Michael Smith and Ranjan Subramanian and Xiaoqing Ellen Tan and Binh Tang and Ross Taylor and Adina Williams and Jian Xiang Kuan and Puxin Xu and Zheng Yan and Iliyan Zarov and Yuchen Zhang and Angela Fan and Melanie Kambadur and Sharan Narang and Aurelien Rodriguez and Robert Stojnic and Sergey Edunov and Thomas Scialom},
      year={2023},
      eprint={2307.09288},
      archivePrefix={arXiv},
      primaryClass={cs.CL}
}

@misc{Belkhale24-rth,
      title={RT-H: Action Hierarchies Using Language}, 
      author={Suneel Belkhale and Tianli Ding and Ted Xiao and Pierre Sermanet and Quon Vuong and Jonathan Tompson and Yevgen Chebotar and Debidatta Dwibedi and Dorsa Sadigh},
      year={2024},
      eprint={2403.01823},
      archivePrefix={arXiv},
      primaryClass={cs.RO}
}

@misc{Nvidia-tensorRT-LLM,
  title = {TensorRT-LLM},
  howpublished = {\url{https://github.com/NVIDIA/TensorRT-LLM?tab=readme-ov-file}},
  author = {NVIDIA},
  year = {2024}
}

@misc{Shi24-yayRobot,
      title={Yell At Your Robot: Improving On-the-Fly from Language Corrections}, 
      author={Lucy Xiaoyang Shi and Zheyuan Hu and Tony Z. Zhao and Archit Sharma and Karl Pertsch and Jianlan Luo and Sergey Levine and Chelsea Finn},
      year={2024},
      eprint={2403.12910},
      archivePrefix={arXiv},
      primaryClass={cs.RO}
}

@misc{Ha23-scalingUpDistillingDown,
      title={Scaling Up and Distilling Down: Language-Guided Robot Skill Acquisition}, 
      author={Huy Ha and Pete Florence and Shuran Song},
      year={2023},
      eprint={2307.14535},
      archivePrefix={arXiv},
      primaryClass={cs.RO}
}

@misc{Beyer24-paligemma,
      title={PaliGemma: A versatile 3B VLM for transfer}, 
      author={Lucas Beyer and Andreas Steiner and André Susano Pinto and Alexander Kolesnikov and Xiao Wang and Daniel Salz and Maxim Neumann and Ibrahim Alabdulmohsin and Michael Tschannen and Emanuele Bugliarello and Thomas Unterthiner and Daniel Keysers and Skanda Koppula and Fangyu Liu and Adam Grycner and Alexey Gritsenko and Neil Houlsby and Manoj Kumar and Keran Rong and Julian Eisenschlos and Rishabh Kabra and Matthias Bauer and Matko Bošnjak and Xi Chen and Matthias Minderer and Paul Voigtlaender and Ioana Bica and Ivana Balazevic and Joan Puigcerver and Pinelopi Papalampidi and Olivier Henaff and Xi Xiong and Radu Soricut and Jeremiah Harmsen and Xiaohua Zhai},
      year={2024},
      eprint={2407.07726},
      archivePrefix={arXiv},
      primaryClass={cs.CV},
      url={https://arxiv.org/abs/2407.07726}, 
}

@misc{Pertsch25-fast,
      title={FAST: Efficient Action Tokenization for Vision-Language-Action Models}, 
      author={Karl Pertsch and Kyle Stachowicz and Brian Ichter and Danny Driess and Suraj Nair and Quan Vuong and Oier Mees and Chelsea Finn and Sergey Levine},
      year={2025},
      eprint={2501.09747},
      archivePrefix={arXiv},
      primaryClass={cs.RO},
      url={https://arxiv.org/abs/2501.09747}, 
}

@misc{Black24-pi0,
      title={$\pi_0$: A Vision-Language-Action Flow Model for General Robot Control}, 
      author={Kevin Black and Noah Brown and Danny Driess and Adnan Esmail and Michael Equi and Chelsea Finn and Niccolo Fusai and Lachy Groom and Karol Hausman and Brian Ichter and Szymon Jakubczak and Tim Jones and Liyiming Ke and Sergey Levine and Adrian Li-Bell and Mohith Mothukuri and Suraj Nair and Karl Pertsch and Lucy Xiaoyang Shi and James Tanner and Quan Vuong and Anna Walling and Haohuan Wang and Ury Zhilinsky},
      year={2024},
      eprint={2410.24164},
      archivePrefix={arXiv},
      primaryClass={cs.LG},
      url={https://arxiv.org/abs/2410.24164}, 
}

@misc{Kim25-openVLA-oft,
      title={Fine-Tuning Vision-Language-Action Models: Optimizing Speed and Success}, 
      author={Moo Jin Kim and Chelsea Finn and Percy Liang},
      year={2025},
      eprint={2502.19645},
      archivePrefix={arXiv},
      primaryClass={cs.RO},
      url={https://arxiv.org/abs/2502.19645}, 
}

@inproceedings{Zawalski24-ecot,
    title={Robotic Control via Embodied Chain-of-Thought Reasoning},
    author={Michał Zawalski and William Chen and Karl Pertsch and Oier Mees and Chelsea Finn and Sergey Levine},
    booktitle={Conference on Robot Learning},
    year={2024}
}

@misc{Belkhale24-miniVLA,
      title={MiniVLA: A Better VLA with a Smaller Footprint}, 
      author={Suneel Belkhale and Dorsa Sadigh},
      year={2024},
      url={https://ai.stanford.edu/blog/minivla/}, 
}

@misc{Liu23-libero,
      title={LIBERO: Benchmarking Knowledge Transfer for Lifelong Robot Learning}, 
      author={Bo Liu and Yifeng Zhu and Chongkai Gao and Yihao Feng and Qiang Liu and Yuke Zhu and Peter Stone},
      year={2023},
      eprint={2306.03310},
      archivePrefix={arXiv},
      primaryClass={cs.AI},
      url={https://arxiv.org/abs/2306.03310}, 
}

@misc{GRT25-geminirobotics,
      title={Gemini Robotics: Bringing AI into the Physical World}, 
      author={{Gemini Robotics Team} and Saminda Abeyruwan and Joshua Ainslie and Jean-Baptiste Alayrac and Montserrat Gonzalez Arenas and Travis Armstrong and Ashwin Balakrishna and Robert Baruch and Maria Bauza and Michiel Blokzijl and Steven Bohez and Konstantinos Bousmalis and Anthony Brohan and Thomas Buschmann and Arunkumar Byravan and Serkan Cabi and Ken Caluwaerts and Federico Casarini and Oscar Chang and Jose Enrique Chen and Xi Chen and Hao-Tien Lewis Chiang and Krzysztof Choromanski and David D'Ambrosio and Sudeep Dasari and Todor Davchev and Coline Devin and Norman Di Palo and Tianli Ding and Adil Dostmohamed and Danny Driess and Yilun Du and Debidatta Dwibedi and Michael Elabd and Claudio Fantacci and Cody Fong and Erik Frey and Chuyuan Fu and Marissa Giustina and Keerthana Gopalakrishnan and Laura Graesser and Leonard Hasenclever and Nicolas Heess and Brandon Hernaez and Alexander Herzog and R. Alex Hofer and Jan Humplik and Atil Iscen and Mithun George Jacob and Deepali Jain and Ryan Julian and Dmitry Kalashnikov and M. Emre Karagozler and Stefani Karp and Chase Kew and Jerad Kirkland and Sean Kirmani and Yuheng Kuang and Thomas Lampe and Antoine Laurens and Isabel Leal and Alex X. Lee and Tsang-Wei Edward Lee and Jacky Liang and Yixin Lin and Sharath Maddineni and Anirudha Majumdar and Assaf Hurwitz Michaely and Robert Moreno and Michael Neunert and Francesco Nori and Carolina Parada and Emilio Parisotto and Peter Pastor and Acorn Pooley and Kanishka Rao and Krista Reymann and Dorsa Sadigh and Stefano Saliceti and Pannag Sanketi and Pierre Sermanet and Dhruv Shah and Mohit Sharma and Kathryn Shea and Charles Shu and Vikas Sindhwani and Sumeet Singh and Radu Soricut and Jost Tobias Springenberg and Rachel Sterneck and Razvan Surdulescu and Jie Tan and Jonathan Tompson and Vincent Vanhoucke and Jake Varley and Grace Vesom and Giulia Vezzani and Oriol Vinyals and Ayzaan Wahid and Stefan Welker and Paul Wohlhart and Fei Xia and Ted Xiao and Annie Xie and Jinyu Xie and Peng Xu and Sichun Xu and Ying Xu and Zhuo Xu and Yuxiang Yang and Rui Yao and Sergey Yaroshenko and Wenhao Yu and Wentao Yuan and Jingwei Zhang and Tingnan Zhang and Allan Zhou and Yuxiang Zhou},
      year={2025},
      eprint={2503.20020},
      archivePrefix={arXiv},
      primaryClass={cs.RO},
      url={https://arxiv.org/abs/2503.20020}, 
}

@misc{Gu23-rttrajectory,
      title={RT-Trajectory: Robotic Task Generalization via Hindsight Trajectory Sketches}, 
      author={Jiayuan Gu and Sean Kirmani and Paul Wohlhart and Yao Lu and Montserrat Gonzalez Arenas and Kanishka Rao and Wenhao Yu and Chuyuan Fu and Keerthana Gopalakrishnan and Zhuo Xu and Priya Sundaresan and Peng Xu and Hao Su and Karol Hausman and Chelsea Finn and Quan Vuong and Ted Xiao},
      year={2023},
      eprint={2311.01977},
      archivePrefix={arXiv},
      primaryClass={cs.RO},
      url={https://arxiv.org/abs/2311.01977}, 
}

@misc{Shi25-hiRobot,
      title={Hi Robot: Open-Ended Instruction Following with Hierarchical Vision-Language-Action Models}, 
      author={Lucy Xiaoyang Shi and Brian Ichter and Michael Equi and Liyiming Ke and Karl Pertsch and Quan Vuong and James Tanner and Anna Walling and Haohuan Wang and Niccolo Fusai and Adrian Li-Bell and Danny Driess and Lachy Groom and Sergey Levine and Chelsea Finn},
      year={2025},
      eprint={2502.19417},
      archivePrefix={arXiv},
      primaryClass={cs.RO},
      url={https://arxiv.org/abs/2502.19417}, 
}

@misc{Hwang24-emma,
      title={EMMA: End-to-End Multimodal Model for Autonomous Driving}, 
      author={Jyh-Jing Hwang and Runsheng Xu and Hubert Lin and Wei-Chih Hung and Jingwei Ji and Kristy Choi and Di Huang and Tong He and Paul Covington and Benjamin Sapp and Yin Zhou and James Guo and Dragomir Anguelov and Mingxing Tan},
      year={2024},
      eprint={2410.23262},
      archivePrefix={arXiv},
      primaryClass={cs.CV},
      url={https://arxiv.org/abs/2410.23262}, 
}

@misc{Chen25-tensorrt-openvla,
      title={TensorRT-OpenVLA}, 
      author={William Chen and Michał Zawalski and Karl Pertsch and Oier Mees and Chelsea Finn and Sergey Levine},
      year={2025},
      url={https://github.com/rail-berkeley/tensorrt-openvla}, 
}

@misc{Deitke24-molmo-pixmo,
      title={Molmo and PixMo: Open Weights and Open Data for State-of-the-Art Vision-Language Models}, 
      author={Matt Deitke and Christopher Clark and Sangho Lee and Rohun Tripathi and Yue Yang and Jae Sung Park and Mohammadreza Salehi and Niklas Muennighoff and Kyle Lo and Luca Soldaini and Jiasen Lu and Taira Anderson and Erin Bransom and Kiana Ehsani and Huong Ngo and YenSung Chen and Ajay Patel and Mark Yatskar and Chris Callison-Burch and Andrew Head and Rose Hendrix and Favyen Bastani and Eli VanderBilt and Nathan Lambert and Yvonne Chou and Arnavi Chheda and Jenna Sparks and Sam Skjonsberg and Michael Schmitz and Aaron Sarnat and Byron Bischoff and Pete Walsh and Chris Newell and Piper Wolters and Tanmay Gupta and Kuo-Hao Zeng and Jon Borchardt and Dirk Groeneveld and Crystal Nam and Sophie Lebrecht and Caitlin Wittlif and Carissa Schoenick and Oscar Michel and Ranjay Krishna and Luca Weihs and Noah A. Smith and Hannaneh Hajishirzi and Ross Girshick and Ali Farhadi and Aniruddha Kembhavi},
      year={2024},
      eprint={2409.17146},
      archivePrefix={arXiv},
      primaryClass={cs.CV},
      url={https://arxiv.org/abs/2409.17146}, 
}

@misc{Chen25-ecot-lite,
        title={Training Strategies for Efficient Embodied Reasoning},
        author={William Chen and Suneel Belkhale and Suvir Mirchandani and Oier Mees and Danny Driess and Karl Pertsch and Sergey Levine},
        journal = {Conference on Robot Learning},
        year={2025},
}

@misc{Pi25-pi05,
    title={$\pi_{0.5}$: a Vision-Language-Action Model with Open-World Generalization},
    author={Physical Intelligence and Kevin Black and Noah Brown and James Darpinian and Karan Dhabalia and Danny Driess and Adnan Esmail and Michael Equi and Chelsea Finn and Niccolo Fusai and Manuel Y. Galliker and Dibya Ghosh and Lachy Groom and Karol Hausman and Brian Ichter and Szymon Jakubczak and Tim Jones and Liyiming Ke and Devin LeBlanc and Sergey Levine and Adrian Li-Bell and Mohith Mothukuri and Suraj Nair and Karl Pertsch and Allen Z. Ren and Lucy Xiaoyang Shi and Laura Smith and Jost Tobias Springenberg and Kyle Stachowicz and James Tanner and Quan Vuong and Homer Walke and Anna Walling and Haohuan Wang and Lili Yu and Ury Zhilinsky},
    year={2025}
}

@misc{Li25-hamsterhierarchicalactionmodels,
      title={HAMSTER: Hierarchical Action Models For Open-World Robot Manipulation}, 
      author={Yi Li and Yuquan Deng and Jesse Zhang and Joel Jang and Marius Memmel and Raymond Yu and Caelan Reed Garrett and Fabio Ramos and Dieter Fox and Anqi Li and Abhishek Gupta and Ankit Goyal},
      year={2025},
      eprint={2502.05485},
      archivePrefix={arXiv},
      primaryClass={cs.RO},
      url={https://arxiv.org/abs/2502.05485}, 
}

@misc{Wagenmaker25-dsrl,
      title={Steering Your Diffusion Policy with Latent Space Reinforcement Learning}, 
      author={Andrew Wagenmaker and Mitsuhiko Nakamoto and Yunchu Zhang and Seohong Park and Waleed Yagoub and Anusha Nagabandi and Abhishek Gupta and Sergey Levine},
      year={2025},
      eprint={2506.15799},
      archivePrefix={arXiv},
      primaryClass={cs.RO},
      url={https://arxiv.org/abs/2506.15799}, 
}

@article{Nakamoto24-vgps,
  author    = {Mitsuhiko Nakamoto and Oier Mees and Aviral Kumar and Sergey Levine},
  title     = {Steering Your Generalists: Improving Robotic Foundation Models via Value Guidance},
  journal   = {Conference on Robot Learning (CoRL)},
  year      = {2024},
}

@misc{Frans25-cfg-rl,
      title={Diffusion Guidance Is a Controllable Policy Improvement Operator}, 
      author={Kevin Frans and Seohong Park and Pieter Abbeel and Sergey Levine},
      year={2025},
      eprint={2505.23458},
      archivePrefix={arXiv},
      primaryClass={cs.LG},
      url={https://arxiv.org/abs/2505.23458}, 
}

@misc{Ho22-cfg,
      title={Classifier-Free Diffusion Guidance}, 
      author={Jonathan Ho and Tim Salimans},
      year={2022},
}

@misc{Dhariwal21-diffusion-classifier-guidance,
      title={Diffusion Models Beat GANs on Image Synthesis}, 
      author={Prafulla Dhariwal and Alex Nichol},
      year={2021},
      eprint={2105.05233},
      archivePrefix={arXiv},
      primaryClass={cs.LG},
      url={https://arxiv.org/abs/2105.05233}, 
}

@inproceedings{Du25-dynaguide,
    title={DynaGuide: Steering Diffusion Policies with Active Dynamic Guidance},
    author={Maximilian Du and Shuran Song},
    booktitle={Proceedings of the 39th Conference on Neural Information Processing Systems (NeurIPS)},
    year={2025}
}

@misc{Wang24-inference-time-policy-steering,
    title={Inference-Time Policy Steering through Human Interactions},
    author={Yanwei Wang and Lirui Wang and Yilun Du and Balakumar Sundaralingam and Xuning Yang and Yu-Wei Chao and Claudia Perez-D'Arpino and Dieter Fox and Julie Shah},
    year={2024}
}

@misc{Li21-prefix-tuning,
      title={Prefix-Tuning: Optimizing Continuous Prompts for Generation}, 
      author={Xiang Lisa Li and Percy Liang},
      year={2021},
      eprint={2101.00190},
      archivePrefix={arXiv},
      primaryClass={cs.CL},
      url={https://arxiv.org/abs/2101.00190}, 
}

@misc{Kwok25-robomonkey,
      title={RoboMonkey: Scaling Test-Time Sampling and Verification for Vision-Language-Action Models}, 
      author={Jacky Kwok and Christopher Agia and Rohan Sinha and Matt Foutter and Shulu Li and Ion Stoica and Azalia Mirhoseini and Marco Pavone},
      year={2025},
      eprint={2506.17811},
      archivePrefix={arXiv},
      primaryClass={cs.RO},
      url={https://arxiv.org/abs/2506.17811}, 
}

@misc{Wu25-forewarn,
		title={From Foresight to Forethought: VLM-In-the-Loop Policy Steering via Latent Alignment}, 
		author={Yilin Wu and Ran Tian and Gokul Swamy and Andrea Bajcsy},
		year={2025},
		eprint={2502.01828},
		archivePrefix={arXiv},
		primaryClass={cs.RO},
		url={https://arxiv.org/abs/2502.01828}, 
  
}

@misc{Glossop25-cast,
      title={CAST: Counterfactual Labels Improve Instruction Following in Vision-Language-Action Models}, 
      author={Catherine Glossop and William Chen and Arjun Bhorkar and Dhruv Shah and Sergey Levine},
      year={2025},
      eprint={2508.13446},
      archivePrefix={arXiv},
      primaryClass={cs.RO},
      url={https://arxiv.org/abs/2508.13446}, 
}

@inproceedings{Betker23-dalle3,
  title={Improving Image Generation with Better Captions},
  author={James Betker and Gabriel Goh and Li Jing and Tim Brooks and Jianfeng Wang and Linjie Li and LongOuyang and Juntang Zhuang and Joyce Lee and Yufei Guo and Wesam Manassra and Prafulla Dhariwal and Casey Chu and Yunxin Jiao and Aditya Ramesh},
  year={2023}
}

@misc{Zhang24-sprint,
      title={SPRINT: Scalable Policy Pre-Training via Language Instruction Relabeling}, 
      author={Jesse Zhang and Karl Pertsch and Jiahui Zhang and Joseph J. Lim},
      year={2024},
      eprint={2306.11886},
      archivePrefix={arXiv},
      primaryClass={cs.RO},
      url={https://arxiv.org/abs/2306.11886}, 
}

@misc{Lynch22-interactiveLanguage,
      title={Interactive Language: Talking to Robots in Real Time}, 
      author={Corey Lynch and Ayzaan Wahid and Jonathan Tompson and Tianli Ding and James Betker and Robert Baruch and Travis Armstrong and Pete Florence},
      year={2022}, 
}

@misc{Xiao22-robotSkillAcquisitionVLM,
      title={Robotic Skill Acquisition via Instruction Augmentation with Vision-Language Models}, 
      author={Ted Xiao and Harris Chan and Pierre Sermanet and Ayzaan Wahid and Anthony Brohan and Karol Hausman and Sergey Levine and Jonathan Tompson},
      year={2022}, 
}

@misc{Zheng25-tracevla,
      title={TraceVLA: Visual Trace Prompting Enhances Spatial-Temporal Awareness for Generalist Robotic Policies}, 
      author={Ruijie Zheng and Yongyuan Liang and Shuaiyi Huang and Jianfeng Gao and Hal Daumé III and Andrey Kolobov and Furong Huang and Jianwei Yang},
      year={2025},
      eprint={2412.10345},
      archivePrefix={arXiv},
      primaryClass={cs.RO},
      url={https://arxiv.org/abs/2412.10345}, 
}

@misc{Smith24-steer,
      title={STEER: Flexible Robotic Manipulation via Dense Language Grounding}, 
      author={Laura Smith and Alex Irpan and Montserrat Gonzalez Arenas and Sean Kirmani and Dmitry Kalashnikov and Dhruv Shah and Ted Xiao},
      year={2024},
      eprint={2411.03409},
      archivePrefix={arXiv},
      primaryClass={cs.RO},
      url={https://arxiv.org/abs/2411.03409}, 
}

@misc{Hirose25-omnivla,
      title={OmniVLA: An Omni-Modal Vision-Language-Action Model for Robot Navigation}, 
      author={Noriaki Hirose and Catherine Glossop and Dhruv Shah and Sergey Levine},
      year={2025},
      eprint={2509.19480},
      archivePrefix={arXiv},
      primaryClass={cs.RO},
      url={https://arxiv.org/abs/2509.19480}, 
}

@misc{Lee25-molmoAct,
      title={MolmoAct: Action Reasoning Models that can Reason in Space}, 
      author={Jason Lee and Jiafei Duan and Haoquan Fang and Yuquan Deng and Shuo Liu and Boyang Li and Bohan Fang and Jieyu Zhang and Yi Ru Wang and Sangho Lee and Winson Han and Wilbert Pumacay and Angelica Wu and Rose Hendrix and Karen Farley and Eli VanderBilt and Ali Farhadi and Dieter Fox and Ranjay Krishna},
      year={2025},
      eprint={2508.07917},
      archivePrefix={arXiv},
      primaryClass={cs.RO},
      url={https://arxiv.org/abs/2508.07917}, 
}

@misc{li2025interactivetaskplanninglanguage,
      title={Interactive Task Planning with Language Models}, 
      author={Boyi Li and Philipp Wu and Pieter Abbeel and Jitendra Malik},
      year={2025},
      eprint={2310.10645},
      archivePrefix={arXiv},
      primaryClass={cs.RO},
      url={https://arxiv.org/abs/2310.10645}, 
}

@misc{shah2022lmnavroboticnavigationlarge,
      title={LM-Nav: Robotic Navigation with Large Pre-Trained Models of Language, Vision, and Action}, 
      author={Dhruv Shah and Blazej Osinski and Brian Ichter and Sergey Levine},
      year={2022},
      eprint={2207.04429},
      archivePrefix={arXiv},
      primaryClass={cs.RO},
      url={https://arxiv.org/abs/2207.04429}, 
}

@misc{Driess25-kiVLA,
      title={Knowledge Insulating Vision-Language-Action Models: Train Fast, Run Fast, Generalize Better}, 
      author={Danny Driess and Jost Tobias Springenberg and Brian Ichter and Lili Yu and Adrian Li-Bell and Karl Pertsch and Allen Z. Ren and Homer Walke and Quan Vuong and Lucy Xiaoyang Shi and Sergey Levine},
      year={2025},
      eprint={2505.23705},
      archivePrefix={arXiv},
      primaryClass={cs.LG},
      url={https://arxiv.org/abs/2505.23705}, 
}

@book{Kahneman11-thinkingFastAndSlow,
  author = {Kahneman, Daniel},
  description = {Thinking, Fast and Slow},
  publisher = {Farrar, Straus and Giroux},
  title = {Thinking, fast and slow},
  year = 2011
}

@misc{Wanna26-limitedLinguisticDiversityVLAs,
      title={Limited Linguistic Diversity in Embodied AI Datasets}, 
      author={Selma Wanna and Agnes Luhtaru and Jonathan Salfity and Ryan Barron and Juston Moore and Cynthia Matuszek and Mitch Pryor},
      year={2026},
      eprint={2601.03136},
      archivePrefix={arXiv},
      primaryClass={cs.CL},
      url={https://arxiv.org/abs/2601.03136}, 
}

@misc{Zhou25-liberoPro,
      title={LIBERO-PRO: Towards Robust and Fair Evaluation of Vision-Language-Action Models Beyond Memorization}, 
      author={Xueyang Zhou and Yangming Xu and Guiyao Tie and Yongchao Chen and Guowen Zhang and Duanfeng Chu and Pan Zhou and Lichao Sun},
      year={2025},
      eprint={2510.03827},
      archivePrefix={arXiv},
      primaryClass={cs.CV},
      url={https://arxiv.org/abs/2510.03827}, 
}

@misc{huang2022languagemodelszeroshotplanners,
      title={Language Models as Zero-Shot Planners: Extracting Actionable Knowledge for Embodied Agents}, 
      author={Wenlong Huang and Pieter Abbeel and Deepak Pathak and Igor Mordatch},
      year={2022},
      eprint={2201.07207},
      archivePrefix={arXiv},
      primaryClass={cs.LG},
      url={https://arxiv.org/abs/2201.07207}, 
}

@misc{yang2025lohovlaunifiedvisionlanguageactionmodel,
      title={LoHoVLA: A Unified Vision-Language-Action Model for Long-Horizon Embodied Tasks}, 
      author={Yi Yang and Jiaxuan Sun and Siqi Kou and Yihan Wang and Zhijie Deng},
      year={2025},
      eprint={2506.00411},
      archivePrefix={arXiv},
      primaryClass={cs.RO},
      url={https://arxiv.org/abs/2506.00411}, 
}

@misc{peschl2025codeactionhierarchicallearning,
      title={From Code to Action: Hierarchical Learning of Diffusion-VLM Policies}, 
      author={Markus Peschl and Pietro Mazzaglia and Daniel Dijkman},
      year={2025},
      eprint={2509.24917},
      archivePrefix={arXiv},
      primaryClass={cs.RO},
      url={https://arxiv.org/abs/2509.24917}, 
}

@misc{fang2025robixunifiedmodelrobot,
      title={Robix: A Unified Model for Robot Interaction, Reasoning and Planning}, 
      author={Huang Fang and Mengxi Zhang and Heng Dong and Wei Li and Zixuan Wang and Qifeng Zhang and Xueyun Tian and Yucheng Hu and Hang Li},
      year={2025},
      eprint={2509.01106},
      archivePrefix={arXiv},
      primaryClass={cs.AI},
      url={https://arxiv.org/abs/2509.01106}, 
}

@misc{zhang2023bootstrapskillslearningsolve,
      title={Bootstrap Your Own Skills: Learning to Solve New Tasks with Large Language Model Guidance}, 
      author={Jesse Zhang and Jiahui Zhang and Karl Pertsch and Ziyi Liu and Xiang Ren and Minsuk Chang and Shao-Hua Sun and Joseph J. Lim},
      year={2023},
      eprint={2310.10021},
      archivePrefix={arXiv},
      primaryClass={cs.RO},
      url={https://arxiv.org/abs/2310.10021}, 
}

@article{Fu24-icrt,
    title={In-Context Imitation Learning via Next-Token Prediction}, 
    author={Letian Fu and Huang Huang and Gaurav Datta and Lawrence Yunliang Chen and William Chung-Ho Panitch and Fangchen Liu and Hui Li and Ken Goldberg},
    journal={arXiv preprint arXiv:2408.15980},
    year={2024}
}

@misc{Yin25-robotInContextLearningLLMs,
      title={In-Context Learning Enables Robot Action Prediction in LLMs}, 
      author={Yida Yin and Zekai Wang and Yuvan Sharma and Dantong Niu and Trevor Darrell and Roei Herzig},
      year={2025},
      eprint={2410.12782},
      archivePrefix={arXiv},
      primaryClass={cs.RO},
      url={https://arxiv.org/abs/2410.12782}, 
}

@misc{jiang2025galaxeaopenworlddatasetg0,
      title={Galaxea Open-World Dataset and G0 Dual-System VLA Model}, 
      author={Tao Jiang and Tianyuan Yuan and Yicheng Liu and Chenhao Lu and Jianning Cui and Xiao Liu and Shuiqi Cheng and Jiyang Gao and Huazhe Xu and Hang Zhao},
      year={2025},
      eprint={2509.00576},
      archivePrefix={arXiv},
      primaryClass={cs.RO},
      url={https://arxiv.org/abs/2509.00576}, 
}

@misc{Carion20-detr,
      title={End-to-End Object Detection with Transformers}, 
      author={Nicolas Carion and Francisco Massa and Gabriel Synnaeve and Nicolas Usunier and Alexander Kirillov and Sergey Zagoruyko},
      year={2020},
      eprint={2005.12872},
      archivePrefix={arXiv},
      primaryClass={cs.CV},
      url={https://arxiv.org/abs/2005.12872}, 
}

@article{Goodman16-rationalSpeechActs,
title = {Pragmatic Language Interpretation as Probabilistic Inference},
journal = {Trends in Cognitive Sciences},
volume = {20},
number = {11},
pages = {818-829},
year = {2016},
issn = {1364-6613},
doi = {https://doi.org/10.1016/j.tics.2016.08.005},
url = {https://www.sciencedirect.com/science/article/pii/S136466131630122X},
author = {Noah D. Goodman and Michael C. Frank}
}
\clearpage
\appendices


\section{Synthetic Generations in Bridge}
\label[appendix]{app-sec:generating-for-bridge}

\subsection{Generating Subtasks and Steering Commands}
\label[appendix]{app-subsec:generating-subtasks-commands}
We present the Gemini~\citep{GeminiTeam24-gemini} prompts for extracting subtasks in \cref{fig:bridge-subtask-prompt}. This decomposes each Bridge episode into subtasks, as \citet{Zawalski24-ecot} did. These subtasks are used in conjunction with bounding boxes, gripper trace coordinates, and motions (discussed below) in the prompt shown in \cref{fig:bridge-steerable-command-prompt} to produce the steering commands used for training our Steerable Policy. The result is that each episode is divided into subtasks, and for each subtask, it has a list of corresponding steering commands (of all the command styles discussed in \cref{subsec:types-of-steerable-commands}).

Note that all pixel coordinates are normalized from 0 to 255, labeled as bracketed tuples (first number represents column from the left and second row from the top, with $[0, 0]$ being the top-right corner).

\subsection{Generating Rationalizations}
\label[appendix]{app-subsec:generating-rationalizatons}
Once we have a subtask decomposition of each trajectory in Bridge, we can produce rationalizations for explaining why each subtask is appropriate for their corresponding starting frames. We do this by again querying Gemini 2.0, giving it each subtask's initial observation, the overall and past subtasks, and the current subtask for it to explain post-hoc. See \cref{fig:bridge-rationale-prompt} for the full prompt. This attaches a rationale for each subtask in each episode.

\subsection{Extracting Embodied Features}
\label[appendix]{app-subsec:embodied-features}
Our pipeline for extracting object and gripper coordinates follows three stages.

First, we identify relevant objects in each scene using the \texttt{MolmoE-1B-0924} model~\citep{Deitke24-molmo-pixmo}. The model is prompted with a natural language description of the robot task and instructed to return a list of up to four objects in the scene, excluding the robot itself. The prompt follows the format: \textit{``The robot task is \{task\_name\}. Briefly provide a list of up to 4 objects in the scene (ignore the robot), including the ones in the task. Don’t add any additional text.''}

Next, we use \texttt{Molmo-7B-D-0924} to predict the coordinates of each identified object in the first image of the trajectory~\citep{Deitke24-molmo-pixmo}. To extend these localizations across time, we apply the Segment Anything Model (SAM 2~\citep{Ravi24-sam2}) to propagate the object points throughout the trajectory, producing temporally consistent segmentation masks.

For gripper coordinates, we manually annotated 100 images from the Bridge dataset with gripper positions. A lightweight DETR model was trained on this data and then used to label gripper centroids across all images in the dataset.

In a final post-processing step, we apply a simple heuristic to filter out task-irrelevant objects by checking whether object names appear in the task description. All object and gripper coordinates are then converted into the \texttt{<loc>} format expected by the PaliGemma model~\citep{Beyer24-paligemma}. Finally, the segmentation masks are tokenized using PaliGemma's vector quantization tokenizer to generate \texttt{<loc><seg>} token sequences.

\section{Details on Hierarchical Methods with In-context Learning Off-the-Shelf Foundation VLMs}
\label[appendix]{app-sec:api-vlms-details}

We now give further details on the in-context learning high-level VLM experiments. All hierarchical methods in this experimental suite are instantiated with Gemini 3.0~\citep{GeminiTeam24-gemini}.

\subsection{Our Approach and Non-reasoning Ablation}
The full VLM prompt for our approach is provided in \cref{fig:gemini-all-prompt}. The VLM is given a task description, an explanation of all steering command styles with in context examples, and a history of past commands and observations. The model is encouraged to reason in-context before providing a final steerable command to instruct our low-level steerable policy. This steerable command is run for $N=20$ steps before the VLM is re-queried.

Note that for grounded steering styles, the policy is allowed to provide keypoints, for example, \robotcmd{move <carrot on left> to <plate>}. Within each angle brackets, the policy may describe a single point in the scene. These descriptions are converted into coordinates, for example, \robotcmd{move $[x_1, y_1]$ to $[x_2, y_2]$} by another Gemini 3.0 call as described in \cref{fig:gemini-pointing-prompt}. In our final approach, we allow the policy to output pointing commands, but not grounded gripper traces, as the VLM performs poorly at providing them.

In our approach, the VLM is instructed to explicitly reason and is additionally allocated a budget of $1024$ thinking tokens through the Gemini API~\citep{GeminiTeam24-gemini}. Our non-reasoning baseline uses the exact same prompt, except that it is told \textit{not} to reason, and is provided a thinking budget of $0$ thinking tokens. We find that the model listens to these instructions and does not reason. Notably, our non-reasoning baseline still has access to an explanation of all steering command styles with in context examples, and a history of past commands and observations to make inferences. The full prompt for our ablation can be found in \cref{fig:gemini-asap-prompt}.

\subsection{SayCan-like Baseline}
The SayCan-like baseline uses the exact same prompt structure as our full approach, with the only change being that it only has an explanation and in-context examples of subtask-level steering commands. The model is still provided with a $1024$ token thinking budget and is told to explicitly reason about the scene and provided history. The full prompt for this baseline is described in \cref{fig:gemini-subtask-prompt}.

\section{Experimental Task Details}
\label[appendix]{app-sec:experiment-details}

\begin{table*}[]
\centering
\resizebox{\textwidth}{!}{%
\begin{tabular}{@{}cccccccccc@{}}
\toprule
\multirow{2}{*}{\textbf{Split}} &
  \multirow{2}{*}{\textbf{Task}} &
  \multicolumn{2}{c}{\textbf{Standard VLAs}} &
  \multicolumn{2}{c}{\textbf{ECoT-Lite Variants}} &
  \multirow{2}{*}{\textbf{\begin{tabular}[c]{@{}c@{}}Full\\ ECoT\end{tabular}}} &
  \multicolumn{3}{c}{\textbf{\begin{tabular}[c]{@{}c@{}}Steerable Policy +\\ Embodied Reasoner (Ours)\end{tabular}}} \\
 &
   &
  \textbf{OpenVLA} &
  \textbf{$\pi_{0.5}$} &
  \textbf{\begin{tabular}[c]{@{}c@{}}Reasoning\\ Dropout\end{tabular}} &
  \textbf{\begin{tabular}[c]{@{}c@{}}Reasoning\\ Pre-training\end{tabular}} &
   &
  \textbf{\begin{tabular}[c]{@{}c@{}}Non-reasoning\\ Ablation\end{tabular}} &
  \textbf{\begin{tabular}[c]{@{}c@{}}Full Approach\\ OpenVLA\end{tabular}} &
  \textbf{\begin{tabular}[c]{@{}c@{}}Full Approach\\ $\pi_{0.5}$\end{tabular}} \\ \midrule
\multirow{2}{*}{\textbf{In Dist.}} &
  \begin{tabular}[c]{@{}c@{}}Put the {[}mushroom / corn / egpplant{]} \\ in the {[}pot / bowl{]}\end{tabular} &
  72.2 &
  61.1 &
  72.2 &
  88.9 &
  88.9 &
  72.2 &
  88.9 &
  \textbf{94.4} \\
 &
  \begin{tabular}[c]{@{}c@{}}Place the {[}spoon / carrot{]} {[}in / on{]}\\ the {[}plate / towel{]}\end{tabular} &
  75.0 &
  \textbf{100} &
  66.7 &
  91.7 &
  91.7 &
  66.7 &
  91.7 &
  83.3 \\ \midrule
\multirow{2}{*}{\textbf{\begin{tabular}[c]{@{}c@{}}In Dist.\\ Challenge\end{tabular}}} &
  \begin{tabular}[c]{@{}c@{}}Put the {[}broccoli / spoon / cube{]} on\\ the towel (all objects are green)\end{tabular} &
  16.7 &
  50.0 &
  66.7 &
  66.7 &
  \textbf{83.3} &
  66.7 &
  66.7 &
  50.0 \\
 &
  \begin{tabular}[c]{@{}c@{}}Put the {[}green / pink{]} spoon\\ on the {[}plate / towel{]}\end{tabular} &
  \textbf{87.5} &
  \textbf{87.5} &
  62.5 &
  37.5 &
  75.0 &
  62.5 &
  75.0 &
  \textbf{87.5} \\ \midrule
\textbf{\begin{tabular}[c]{@{}c@{}}Motion\\ Gen.\end{tabular}} &
  \begin{tabular}[c]{@{}c@{}}Put the {[}carrot / mushroom{]} {[}on / in{]}\\ the {[}plate / pot{]} (target is high up)\end{tabular} &
  8.3 &
  41.6 &
  25.0 &
  33.3 &
  58.3 &
  \textbf{83.3} &
  \textbf{83.3} &
  75.0 \\ \midrule
\multirow{3}{*}{\textbf{\begin{tabular}[c]{@{}c@{}}Spatial\\ Gen.\end{tabular}}} &
  \begin{tabular}[c]{@{}c@{}}Put the {[}banana / tomato{]} in the\\ {[}left / right{]} bowl\end{tabular} &
  91.7 &
  83.3 &
  75.0 &
  83.3 &
  91.7 &
  91.7 &
  \textbf{100} &
  91.7 \\
 &
  \begin{tabular}[c]{@{}c@{}}Put the {[}green / orange{]} toy in the\\ {[}left / right{]} pot\end{tabular} &
  50.0 &
  75.0 &
  75.0 &
  \textbf{87.5} &
  75.0 &
  62.5 &
  \textbf{87.5} &
  \textbf{87.5} \\
 &
  \begin{tabular}[c]{@{}c@{}}Move the {[}mushroom / carrot{]} to the\\ {[}left / right{]} of the {[}other object{]}\end{tabular} &
  62.5 &
  62.5 &
  \textbf{100} &
  \textbf{100} &
  75.0 &
  75.0 &
  87.5 &
  \textbf{100} \\ \midrule
\multirow{4}{*}{\textbf{\begin{tabular}[c]{@{}c@{}}Semantic\\ Gen.\end{tabular}}} &
  \begin{tabular}[c]{@{}c@{}}Put the watermelon\\ on the towel\end{tabular} &
  0 &
  33.3 &
  \textbf{83.3} &
  \textbf{83.3} &
  66.7 &
  50.0 &
  66.7 &
  \textbf{83.3} \\
 &
  \begin{tabular}[c]{@{}c@{}}Put the toothbrush\\ on the plate\end{tabular} &
  33.3 &
  50.0 &
  50.0 &
  83.3 &
  66.7 &
  50.0 &
  \textbf{100} &
  \textbf{100} \\
 &
  \begin{tabular}[c]{@{}c@{}}Put the screw\\ in the bowl\end{tabular} &
  33.3 &
  16.7 &
  16.7 &
  66.7 &
  66.7 &
  50.0 &
  \textbf{83.3} &
  \textbf{83.3} \\
 &
  \begin{tabular}[c]{@{}c@{}}Reach for the\\ {[}ketchup / wrench / mallet{]}\end{tabular} &
  11.1 &
  33.3 &
  22.2 &
  0 &
  \textbf{66.7} &
  33.3 &
  \textbf{66.7} &
  55.5 \\ \midrule
\multicolumn{2}{c}{\textbf{Aggregate}} &
  50.5$\pm$4.7 &
  61.3$\pm$4.6 &
  60.4$\pm$4.6 &
  69.4$\pm$4.4 &
  77.5$\pm$4.0 &
  66.7$\pm$4.5 &
  \textbf{84.6$\pm$3.4} &
  83.7$\pm$3.4 \\ \bottomrule
\end{tabular}%
}
\caption{Full numerical results of the embodied reasoning methods on the task suite defined by \citet{Chen25-ecot-lite}. Numbers represent success percentage, $\pm$StdErr. Equivalent to the data in \cref{fig:reasoner-results}.}
\label{tab:embodied-reasoner-results}
\end{table*}

\begin{table*}[]
\centering
\resizebox{\textwidth}{!}{%
\begin{tabular}{@{}cccccc@{}}
\toprule
\multirow{2}{*}{\textbf{Split}} &
  \multirow{2}{*}{\textbf{Task}} &
  \multirow{2}{*}{\textbf{\begin{tabular}[c]{@{}c@{}}Standard\\ OpenVLA\end{tabular}}} &
  \multirow{2}{*}{\textbf{SayCan-Like}} &
  \multicolumn{2}{c}{\textbf{\begin{tabular}[c]{@{}c@{}}Steerable Policy +\\ In-context Learning VLM(Ours)\end{tabular}}} \\
 &
   &
   &
   &
  \textbf{\begin{tabular}[c]{@{}c@{}}Non-reasoning\\ Ablation\end{tabular}} &
  \textbf{\begin{tabular}[c]{@{}c@{}}Full\\ Approach\end{tabular}} \\ \midrule
\multirow{2}{*}{\textbf{\begin{tabular}[c]{@{}c@{}}Long\\ Horizon 1\end{tabular}}} &
  \begin{tabular}[c]{@{}c@{}}Make the mushroom the\\ only obj. on the plate\end{tabular} &
  35.0 &
  70.0 &
  80.0 &
  100 \\
 &
  \begin{tabular}[c]{@{}c@{}}Make the blue block \\ the only obj. on the plate\end{tabular} &
  30.0 &
  65.0 &
  90.0 &
  80.0 \\ \midrule
\multirow{2}{*}{\textbf{\begin{tabular}[c]{@{}c@{}}Long\\ Horizon 2\end{tabular}}} &
  \begin{tabular}[c]{@{}c@{}}Put all the food in the blue pot \\ and stuffed toys in the tan pot\end{tabular} &
  50.0 &
  36.7 &
  53.3 &
  66.7 \\
 &
  \begin{tabular}[c]{@{}c@{}}Put all the food in the white bowl \\ and stuffed animals in the green bowl\end{tabular} &
  43.3 &
  63.3 &
  80.0 &
  76.7 \\ \midrule
\multirow{2}{*}{\textbf{\begin{tabular}[c]{@{}c@{}}Motion\\ Gen.\end{tabular}}} &
  Stack the pots &
  45.0 &
  85.0 &
  55.0 &
  95.0 \\
 &
  Stack the pots on the towel &
  40.0 &
  60.0 &
  90.0 &
  65.0 \\ \midrule
\multirow{2}{*}{\textbf{\begin{tabular}[c]{@{}c@{}}Spatial\\ Gen.\end{tabular}}} &
  \begin{tabular}[c]{@{}c@{}}Put the banana on the\\ {[}left / right{]} on the plate\end{tabular} &
  60.0 &
  80.0 &
  90.0 &
  100 \\
 &
  \begin{tabular}[c]{@{}c@{}}Put the white and blue stuffed animal\\ in the bowl on the {[}left / right{]}\end{tabular} &
  80.0 &
  90.0 &
  100 &
  100 \\ \midrule
\multirow{2}{*}{\textbf{\begin{tabular}[c]{@{}c@{}}Semantic\\ Gen.\end{tabular}}} &
  Put the hammer on the towel &
  40.0 &
  60.0 &
  70.0 &
  80.0 \\
 &
  Put the screw on the plate &
  60.0 &
  30.0 &
  70.0 &
  80.0 \\ \midrule
\multicolumn{2}{c}{\textbf{Aggregate}} &
  48$\pm$5.1 &
  64$\pm$5.5 &
  78$\pm$4.0 &
  84$\pm$3.7 \\ \bottomrule
\end{tabular}%
}
\caption{Full numerical results of the in-context learning methods on our new multi-step task suite. Numbers represent average task progression (graded with a rubric that decomposes each task into steps), $\pm$StdErr. Equivalent to the data in \cref{fig:multiabstraction-results}.}
\label{tab:long-horizon}
\end{table*}

\begin{table*}[]
\centering
\resizebox{\textwidth}{!}{%
\begin{tabular}{@{}lllc@{}}
\toprule
\textbf{Task} & \textbf{Prompts} & \textbf{Rubric} & \textbf{Success / Fail} \\ \midrule
\multirow{4}{*}{\textbf{Long Horizon 1}} & 1. Make the mushroom the only object on the plate & 1. Interacted with object on plate & \multirow{4}{*}{-} \\
 & 2. Make the blue block the only object on the plate & 2. Plate object no longer on plate &  \\
 & & 3. Pick up correct object &  \\
 & & 4. Put down correct object in correct location &  \\ \midrule
\multirow{6}{*}{\textbf{Long Horizon 2}} & 1. Put all food in blue pot and stuffed animals in tan pot & 1. Pick up object 1 & \multirow{6}{*}{-} \\
 & 2. Put all food in white bowl and stuffed animals in green bowl & 2. Put object 1 in correct container &  \\
 & & 3. Pick up object 2 &  \\
 & & 4. Put object 2 in correct container &  \\
 & & 5. Pick up object 3 &  \\
 & & 6. Put object 3 in correct container &  \\ \midrule
\multirow{4}{*}{\textbf{Motion Gen.}} & 1. Stack the pots & 1. Pick up pot 1 & \multirow{4}{*}{-} \\
 & 2. Stack the pots on the towel & 2. Stack 1 complete &  \\
 & & 3. Pick up pot 2 &  \\
 & & 4. Stack 2 complete &  \\ \midrule
\multirow{2}{*}{\textbf{Spatial Gen.}} & 1. Put the banana on the [right/left] on the plate & 1. Pick up correct object & \multirow{2}{*}{-} \\
 & 2. Put the white and blue stuffed animal in the [left/right] bowl & 2. Put down correct object in correct location &  \\ \midrule
\multirow{2}{*}{\textbf{Semantic Gen.}} & 1. Move the hammer to the towel & 1. Pick up correct object & \multirow{2}{*}{-} \\
 & 2. Move the screw to the plate & 2. Put down correct object in correct location &  \\ \bottomrule
\end{tabular}%
}
\caption{Evaluation Rubric for In-context Learning VLM Experiments. Trials for each task are divided between two prompts. Each trial is scored by a multi-step rubric, and each step is graded by success/fail. Task progress is the average number of successes in a trial. For spatial generalization tasks, we do an equal number of trials with either ``left" or ``right" in the prompt.}
\label{tab:in-context-ruberic}
\end{table*}

The experiments in \cref{sec:experiments} are broadly divided into several generalization splits:
\begin{enumerate}
    \item In-distribution: ``standard'' tasks, akin to those in Bridge~\citep{Ebert21-bridgeDataV1, Walke23-bridgeDataV2}. We expect a VLA trained on Bridge to be proficient at these.
    \item Motion Generalization: Involves standard tasks but with non-standard poses (e.g., placing objects high up).
    \item Spatial Generalization: Tasks involving spatial relation (e.g., ``place in the pot on the left'').
    \item Semantic Generalization: Involves out-of-distribution objects, whose names do not appear in the Bridge dataset.
\end{enumerate}
These splits are used for the human oracle experiments and the embodied reasoner experiments. The off-the-shelf VLM experiments instead use two sets of long-horizon tasks in lieu of the in-distribution ones. In total, we run over 650 robot rollout episodes across our three evaluation suites.

\subsection{Human Oracle Experiments}
\label[appendix]{app-subsec:experiments-human-oracle}

\begin{figure*}
    \centering
    \includegraphics[width=0.8\linewidth]{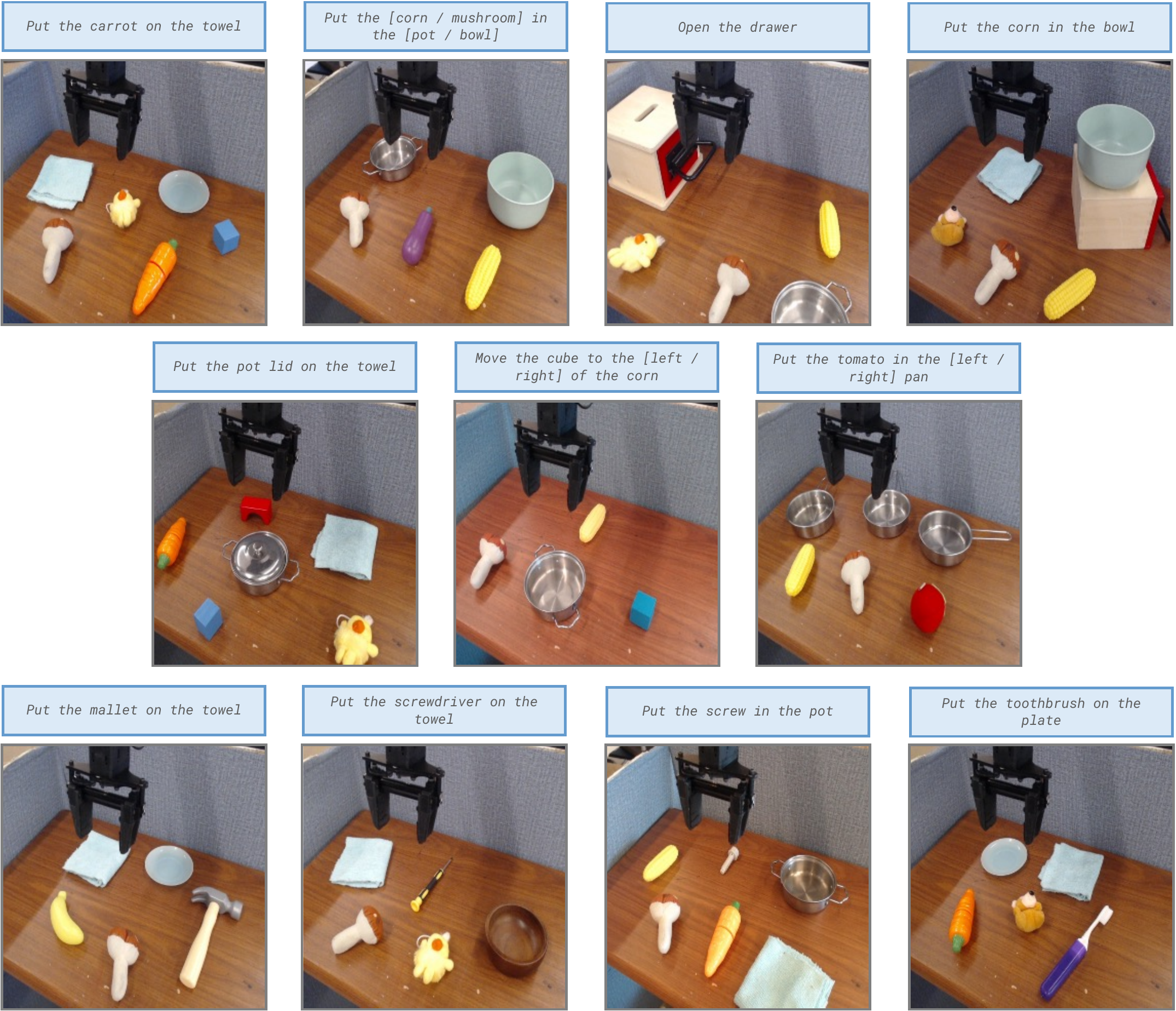}
    \caption{\textbf{Example starting states for the tasks for the didactic experiment} wherein a human operator acts as the high-level policy.}
    \label{fig:oracle-states}
\end{figure*}

The human oracle experiments use the same task splits as the embodied reasoner experiments discussed below. However, they involve different sets of actual tasks, so as to evaluate oracle steering capabilities on a wider range of settings. See \cref{fig:oracle-states} for example initial states.

\subsection{Embodied Reasoner Experiments}
\label[appendix]{app-subsec:experiments-embodied-reasoner}

The tasks for evaluating our hierarchical embodied reasoner are the same as those used for evaluating ECoT-Lite~\citep{Chen25-ecot-lite}. This allows us to compare against other methods for leveraging embodied reasoning data for training VLAs. See \cref{fig:reasoner-states} for example initial states and \cref{tab:embodied-reasoner-results} for detailed task performance.

\subsection{In-context Learning VLM Experiments}
\label[appendix]{app-subsec:experiments-off-the-shelf}
For the in-context learning VLM experiments, we change out all tasks to include more multi-step ones, as they are more suited to evaluating hierarchical systems' generalization. We replace the in-distribution split with two long-horizon tasks, which involve composing many in-distribution behaviors. Additionally, our motion generalization tasks now involves a novel task behavior, though again requires challenging (multi-step) motions to solve.

\begin{figure*}
    \centering
    \includegraphics[width=1\linewidth]{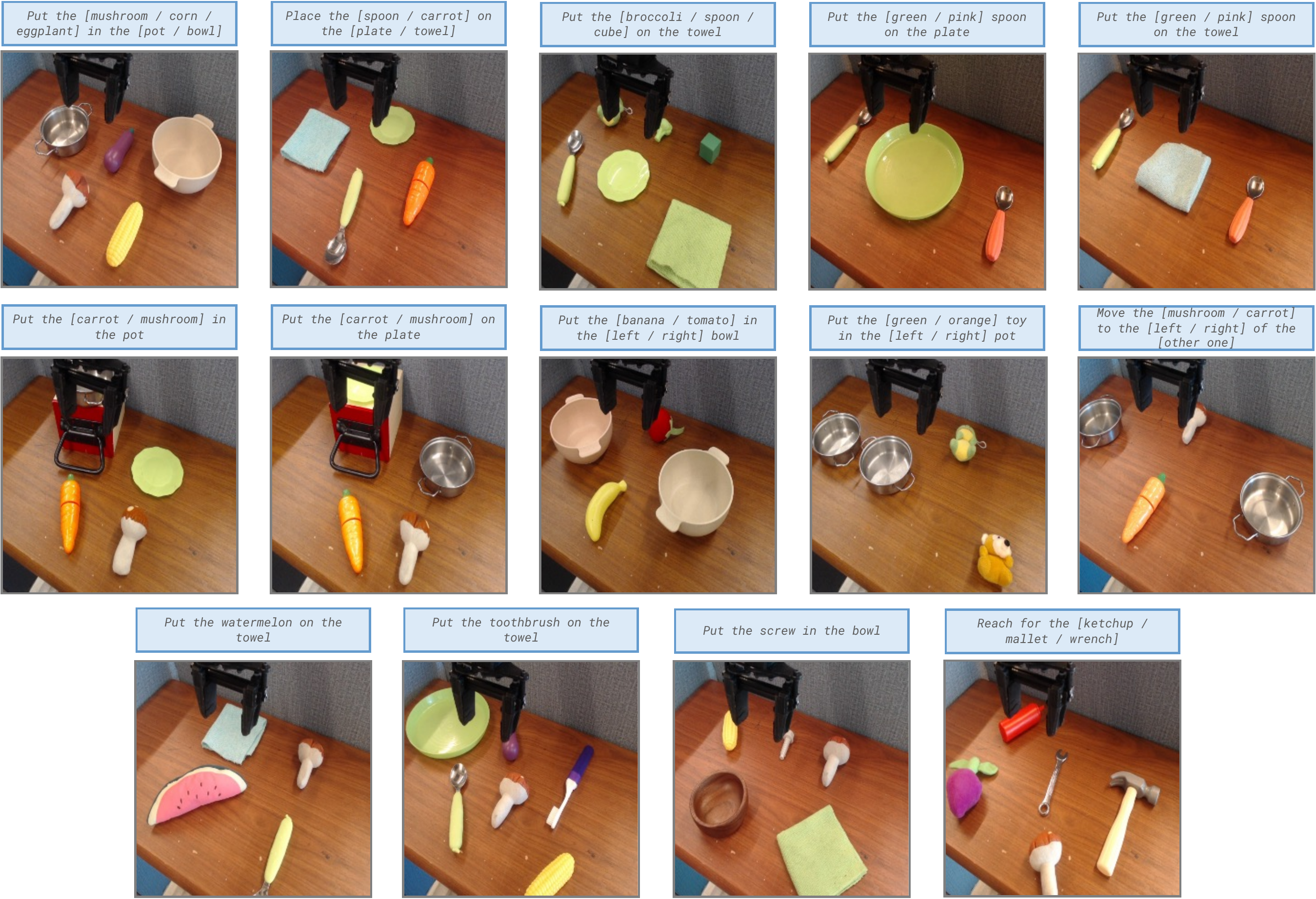}
    \caption{\textbf{Example starting states for the tasks for evaluating embodied reasoning VLAs}, reproduced with permission from~\citet{Chen25-ecot-lite} (as we reuse their task suite).}
    \label{fig:reasoner-states}

    \centering
    \includegraphics[width=1\linewidth]{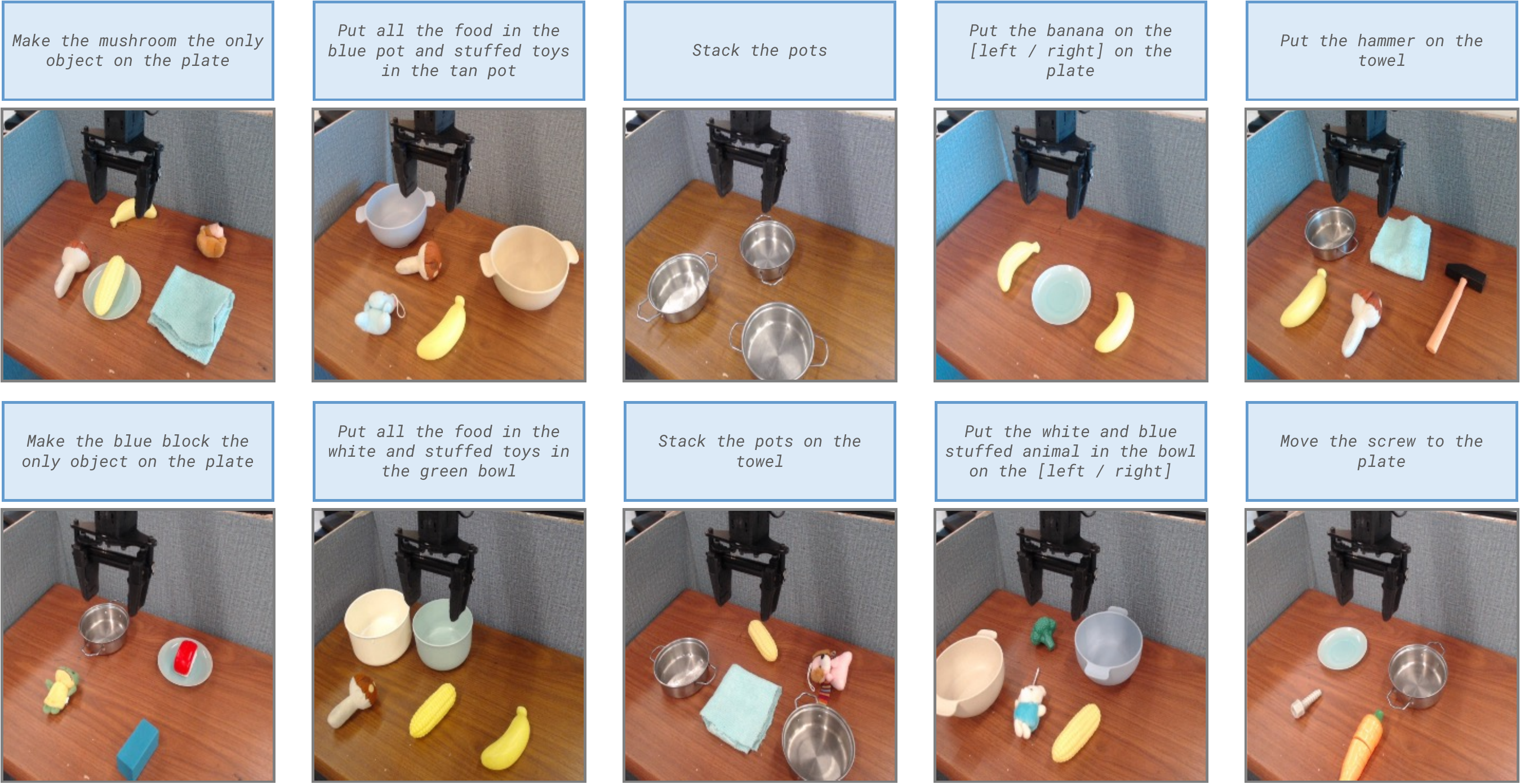}
    \caption{\textbf{Example starting states for the multi-step tasks} for evaluating in-context learning high-level VLMs.}
    \label{fig:icl-states}
\end{figure*}

See \cref{fig:icl-states} for example initial states and \cref{tab:long-horizon} for full results. 
Each trial of each task is graded based on the rubric in \cref{tab:in-context-ruberic}. Each task has two prompts, and trials for each task are divided equally among prompts. Rubrics are graded based on success/fail, and the average number of successes per trial is the final task progress number we report. The policy being evaluated does not need to complete rubric items in order. However, it is penalized for undoing task progress: e.g., the policy may receive credit for putting an object in the correct location, but that credit is revoked if the policy later removes the object. The only exception to this rule is that the policy always gets and maintains credit for interacting with or picking up an object for the first time. The policy is run for $20$ high-level steps ($25$ for Long Horizon 2), and is only terminated early only if all entries in rubric are completed correctly.

\section{Additional Illustrative Examples}
\label[appendix]{app-sec:illustrative-examples}

\subsection{Example Steering Commands}
\label[appendix]{app-subsec:example-steering-commands}

\begin{figure}[h]
    \centering
    \includegraphics[width=\linewidth]{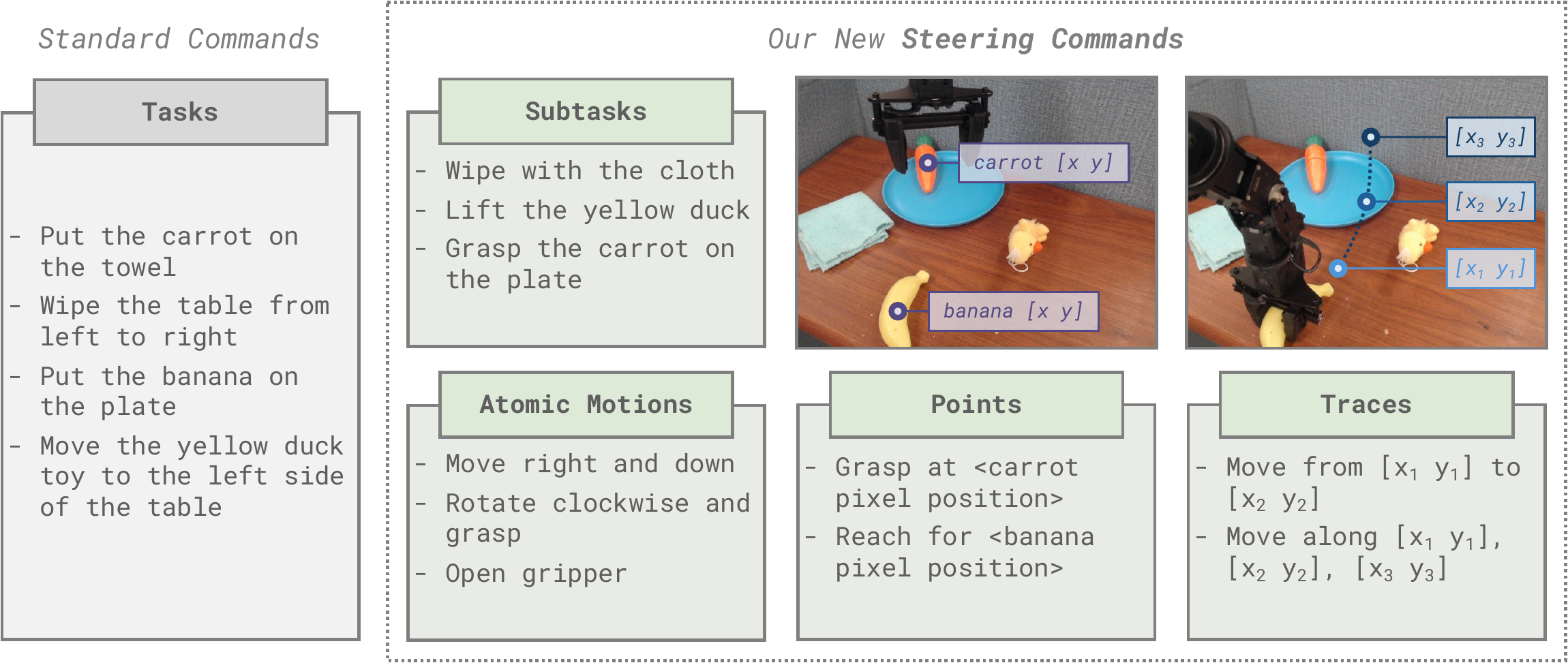}
    \includegraphics[width=\linewidth]{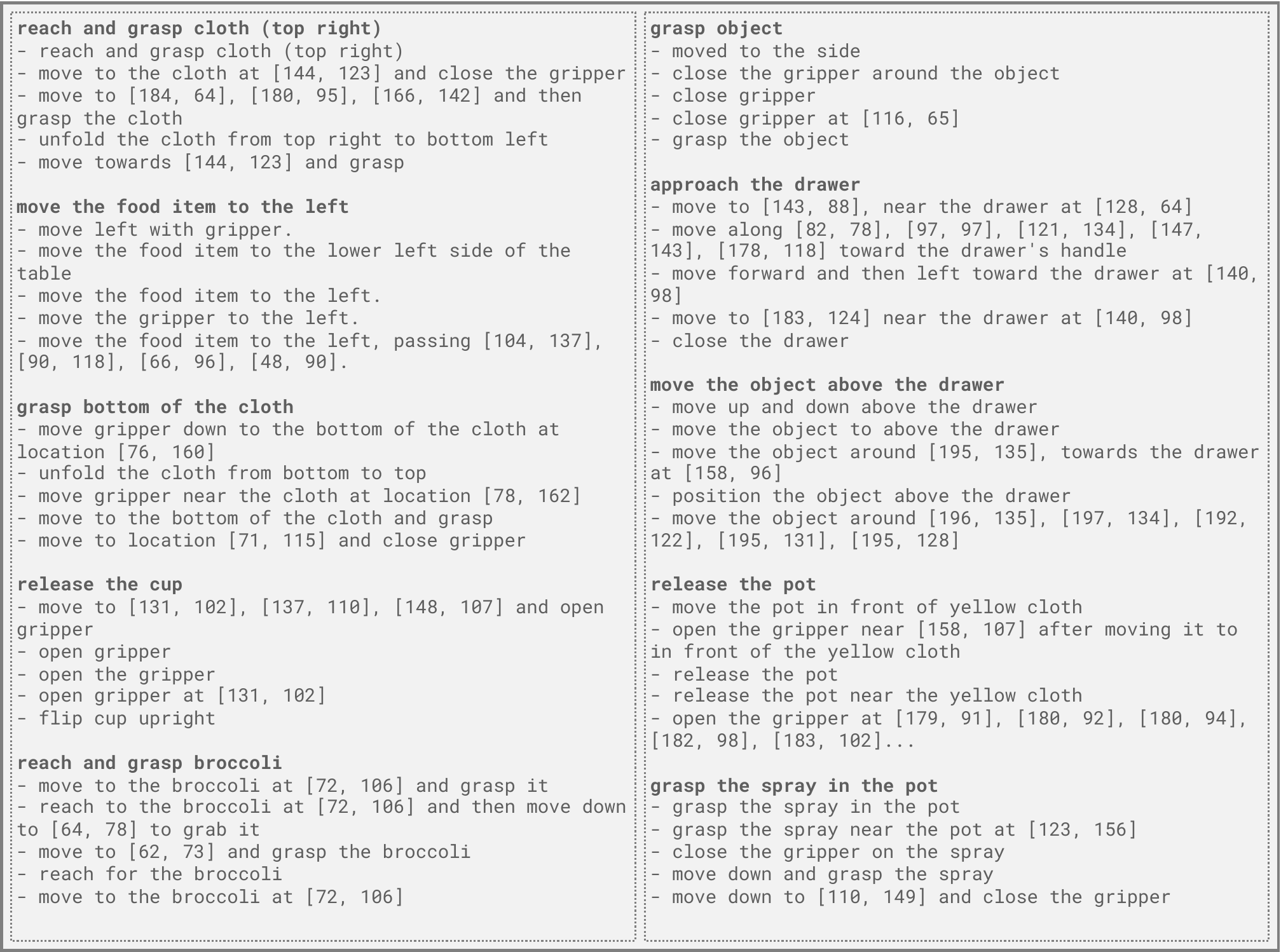}
    \caption{
        \textbf{Example steering commands}. All labels and points in the image are purely for visualization, and do not appear on the actual robot training data images. The list on the bottom is exactly extracted from our training dataset, with bold representing the subtasks and the dashed list representing the corresponding diverse steering commands. Each subtask typically has more than 5 corresponding steering commands, but we limit to that many for readability.
    }
    \label{fig:example-steering-commands}
\end{figure}

We include numerous example steering commands in \cref{fig:example-steering-commands}. 
The diagram is just for visualization purposes, while the text examples are taken exactly from our training data.

\subsection{Example Full Reasoning Traces for High-level Embodied Reasoner}

\begin{figure*}
    \centering
    \includegraphics[width=1\linewidth]{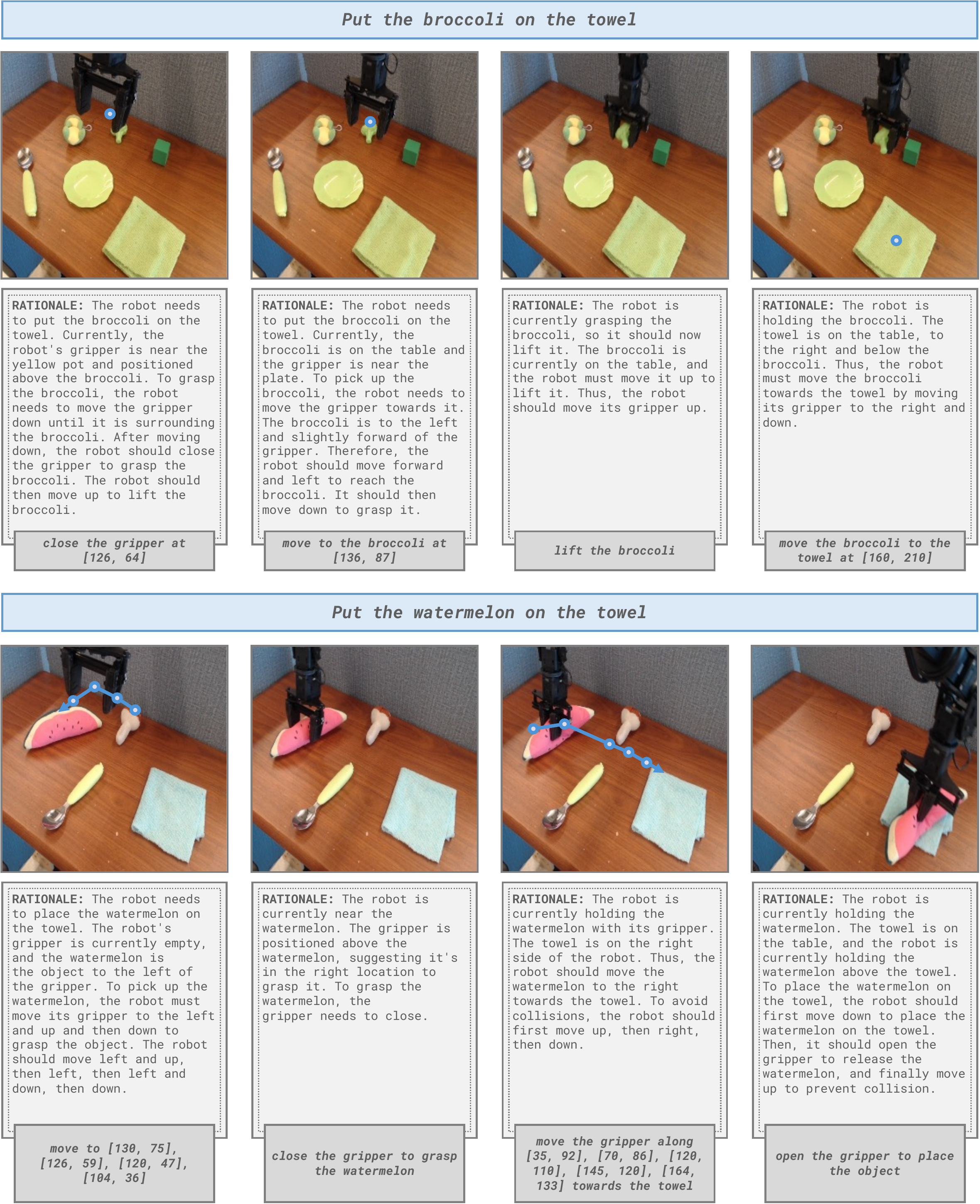}
    \caption{Examples of \textbf{embodied reasonings produced by our fine-tuned VLM}, taken from rollouts for the evaluations in \cref{subsec:results-learned-embodied-reasoners}.}
    \label{fig:example-embodied-reasonings}
\end{figure*}

See \cref{fig:example-embodied-reasonings} for examples of reasoning traces produced by our embodied reasoning high-level VLM.

\subsection{Example Full Reasoning Traces for In-context Learning VLM Methods}
\begin{figure*}
    \centering
    \includegraphics[width=1\linewidth]{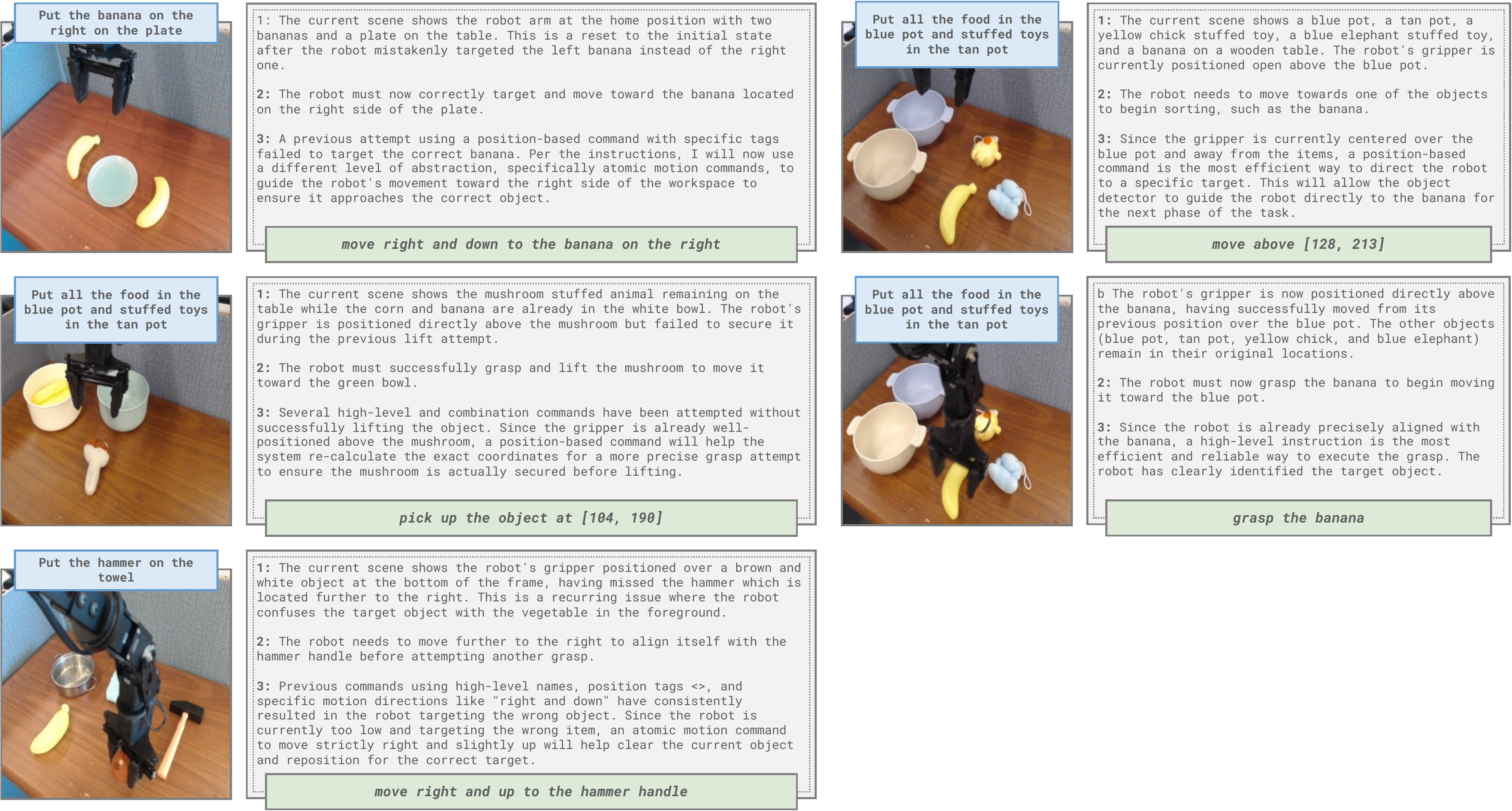}
    \caption{Examples of \textbf{in-context reasonings produced by our off-the-shelf VLM for issuing steering commands}, taken from rollouts for the evaluations in \cref{subsec:results-api-vlms}. These are the unparaphrased versions of the examples in \cref{fig:vlm-examples}}
    \label{fig:example-icl-reasonings}
\end{figure*}

We present the example reasoning traces from our in-context learning VLM method (\cref{subsec:method-api-vlms}) in \cref{fig:example-icl-reasonings}. These are the full, unparaphrased versions of the examples in \cref{fig:vlm-examples}.

For full examples of the equivalent reasoning traces from the SayCan-like baseline, see \cref{fig:subtask-failures} and the associated discussion in \cref{app-subsec:subtasks-insufficient}.

\section{Further Discussions}
\label[appendix]{app-sec:further-discussions}

\subsection{Steering in Robot Learning}

We take \textit{steering or guidance} to mean the broad class of techniques for controlling the outputs of a generative model. As discussed in \cref{sec:related-works}, we specifically consider the case of improving steerability by altering the training strategy. This contrasts with a broad set of works that aim to do steering at inference time, or otherwise without modifying the weights of some pretrained policy. For completeness, we briefly discuss them here.

These approaches usually aim to produce samples that are in-distribution while optimizing some criteria such as a classifier score~\citep{Dhariwal21-diffusion-classifier-guidance}, the likelihood under an unconditional distribution~\cite{Ho22-cfg}, the similarity to a reference output~\citep{Wang24-inference-time-policy-steering}, or a Q-function~\citep{Nakamoto24-vgps, Wagenmaker25-dsrl, Frans25-cfg-rl}. The method for optimizing these criteria also varies, including linearly mixing denoising vector fields~\citep{Dhariwal21-diffusion-classifier-guidance, Ho22-cfg}, optimizing input noise or embeddings~\citep{Wagenmaker25-dsrl, Li21-prefix-tuning, Wang24-inference-time-policy-steering}, or parallelizable sample-and-rank through a scoring function or dynamics model~\citep{Wang24-inference-time-policy-steering, Nakamoto24-vgps, Kwok25-robomonkey, Du25-dynaguide, Wu25-forewarn}.

\subsection{The Manifold of ``Reasonable'' Actions}
\label[appendix]{app-subsec:manifold-of-reasonable-actions}

\begin{figure}
    \centering
    \includegraphics[width=1\linewidth]{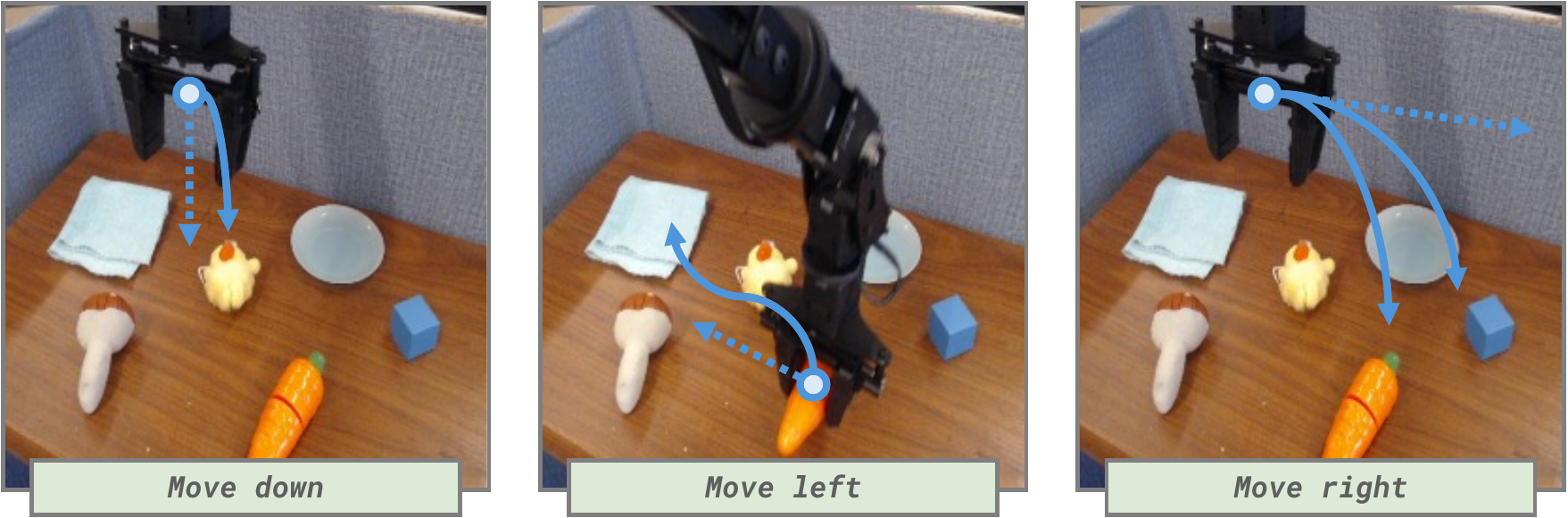}
    \caption{Examples of what we deem the manifold of ``reasonable'' actions, where underspecified commands (especially atomic motions) induce specific intelligent behaviors, based on the observed state. Solid arrows are these ``reasonable'' actions, while the dotted arrow is what the command would look like without conditioning on the image (i.e., akin to controlling the end effector with a game pad and holding a single direction down).
    \textbf{Left:} Rather than moving straight down, the empty gripper would move down towards an object it would grasp. \textbf{Middle:} once an object is grasped, telling the robot to move left causes it to infer where to place it (in the case, the towel, which is left of the gripper). \textbf{Right:} However, this can be ineffective if there are multiple reasonable behaviors. If the robot is commanded to \robotcmd{move right}, it could reach for the carrot \textit{or} the cube.}
    \label{fig:manifold-examples}
\end{figure}

Throughout our experiments, we notice a prevalent qualitative trend in how our Steerable Policy responds to underspecified commands, especially atomic motions. When given such a command, the robot does not \textit{just} move in that direction, but does so while heavily considering the state of the robot and scene. For example, when given the (rather underspecified) command \robotcmd{move left}, it will typically move left \textit{and down}, towards objects that it can interact with. If the gripper is already grasping something, the robot might instead move left \textit{and up}, towards a container that the object can be placed in. Finally, if told ``open gripper'' while near a container, the robot may first move the gripper above it before releasing. See \cref{fig:manifold-examples} for illustrations.

In other words, atomic motion commands seem to cause the policy to sample from the distribution of actions aligned with the commanded direction, but marginalized over all ``reasonable'' tasks that involve moving that way. The VLA can estimate what tasks can be done in a particular scene by attending to the visual observation. Then, when given a motion (e.g., \robotcmd{move left}), it can estimate which of those tasks are likely accomplished by moving left, and then take actions aligned with one or more of those tasks. Intuitively, this arises from common correlations in the training data: if the gripper is empty and a possible steering command for that frame is to \robotcmd{move left}, then the Bridge episode likely involves moving left and down to grasp something. We informally describe this as sticking to the manifold of ``reasonable'' actions.

The qualitative implication is that giving an atomic motion command often results in intelligent motions, rather than ``blindly'' moving in the commanded way -- explaining how atomic motions end up being the best single steering modality in the human oracle experiments from \cref{subsec:oracle-commands}, despite motions being vague \textit{and} the human only being able to issue new commands every two seconds. 

Naturally, this can be a blessing or a curse. While \robotcmd{move left} might cause the robot to pick up an object of interest, it can also get ``confused'' when there are multiple objects to the left, as that command could reasonably cause the policy to pick up either one. Of course, other steering command styles are able to supplement these behaviors, further motivating both training on many steering command styles and applying VLM capabilities to intuit when each style is most appropriate, as we propose.

\subsection{Why are Subtask Commands Insufficient?}
\label[appendix]{app-subsec:subtasks-insufficient}

\begin{figure*}[t]
    \centering
    \includegraphics[width=1\linewidth]{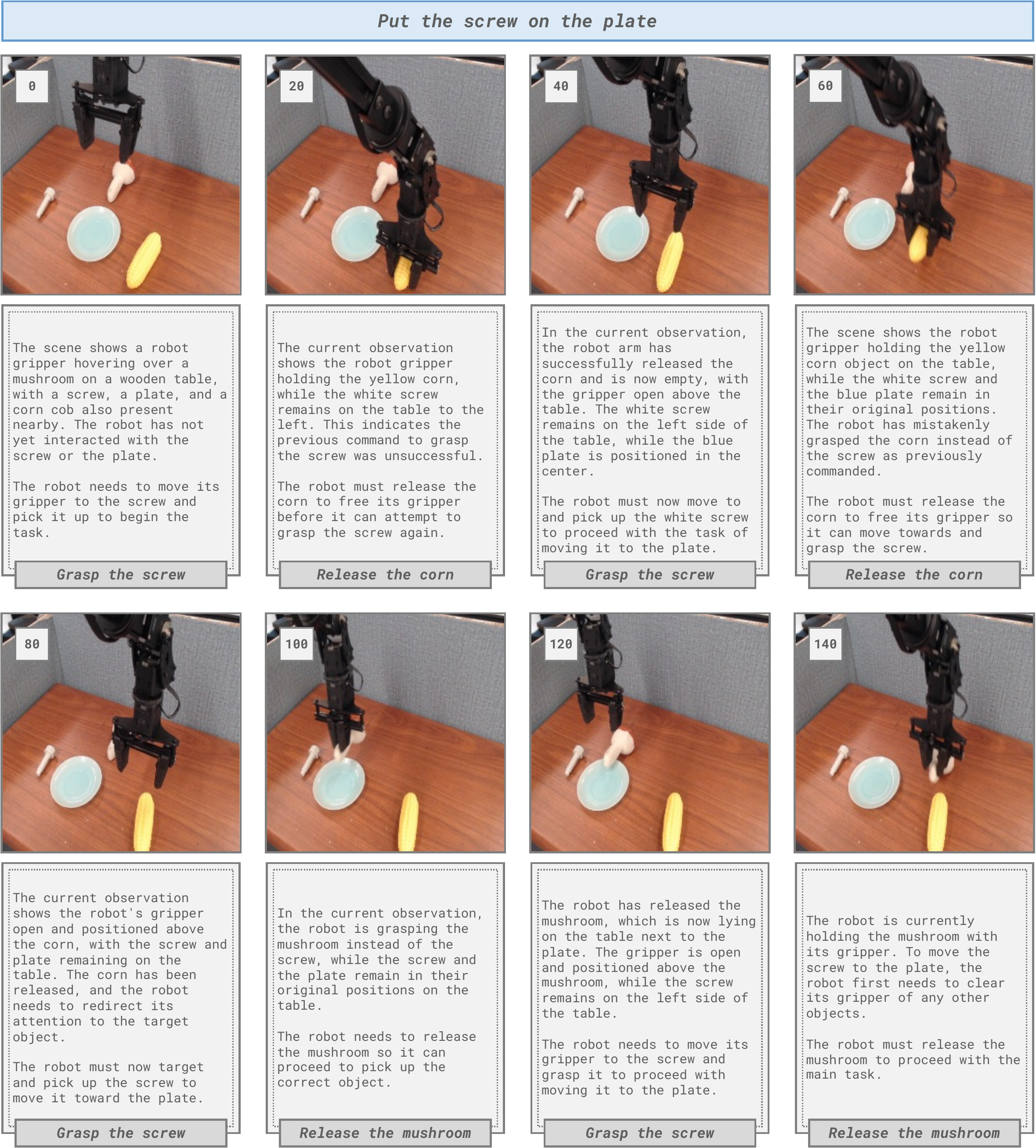}
    \caption{Illustrative example of \textbf{why VLAs cannot leverage VLMs' in-context learning, visual understanding, and reasoning well when issued only subtask commands} (from the SayCan-like baseline in \cref{subsec:results-api-vlms}). The high-level VLM reliably detects incorrect behaviors and what the robot should do to progress the task. \textbf{However, the low-level VLA fails to make use of this when only commanded with subtasks.} Instead, when the VLM correctly tells it to \robotcmd{grasp the screw} (which is out-of-distribution), it always mistakes some other object for it, and grasps the mushroom or corn instead. The policy gets stuck in this loop until the episode times out. Due to lack of steerability with subtasks alone, the high-level VLM has no way to fix this behavior. Our Steerable Policies can get unstuck by simply being commanded at some other level of abstraction.}
    \label{fig:subtask-failures}
\end{figure*}

Standard hierarchical approaches command the low-level policy with tasks and subtasks alone~\citep{Ahn22-sayCan}. However, these steering modalities are too unspecific to reliably induce generalizable behaviors. We show an example of this in \cref{fig:subtask-failures}, drawn directly from our multi-step evaluation tasks. While the high-level VLM knows what the robot should do, subtasks alone are not a viable interface to ``convey'' this behavior to the low-level VLA. Instead, every time the VLM emits \robotcmd{grasp the screw}, the VLA fails to recognize what the screw is, and grasps some distractor object instead. 

While the VLM can tell the policy to drop that incorrect object, it cannot reliably fix this behavior, as subtasks are not flexible enough. This motivates our introduction of diverse steering command -- if one command style fails, then using a different steering modalitiy might work instead (in this case, intuitively telling the Steerable Policy to \robotcmd{move up and left} instead of \robotcmd{grasp the screw} would likely work).

\subsection{What are the Failure Modes of Our Method?}
\label[appendix]{app-subsec:steerability-failures}

Most of the failure modes of our method stem from the \textit{high-level VLM's limited grasp of the low-level Steerable Policy's affordances.} For example, while VLMs can use in-context learning to learn the affordances for the experiments in \cref{subsec:results-api-vlms}, it fails in subtler cases: e.g., the high-level VLM might say to execute the \textit{correct} next step in a multi-step task, but the VLA sometimes fails to follow the command at the current state (e.g., due to dataset bias), and \textit{undoes progress by grasping the same object it just placed}. Instead, the VLA should be prompted with other abstractions, e.g., motions stating to move away \textit{without grasping}. This strategy is hard for VLMs to discover, despite otherwise working well. 

This is effectively an issue with the \textit{pragmatics} of the steering command~\citep{Goodman16-rationalSpeechActs}: the high-level policy must infer the command (and abstraction) that is most likely to communicate and induce a desired behavior in the low-level VLA, selecting from all ``reasonable'' commands for doing so.

Finally, the steerability of our VLA comes from the diversity of behaviors within its data. Training on steering commands simply allows these skills to be induced with the right command style. Conversely, ``narrow'' datasets are \textit{not} conducive to training Steerable Policies. For example, LIBERO~\citep{Liu23-libero} is \textit{not} very diverse: provided trajectories are unimodal~\citep{Zhou25-liberoPro} and starting states for each task are minimally randomized; even minor perturbations degrade performance~\citep{Chen25-ecot-lite}. While policy steering in LIBERO can optimize a narrow behavior prior~\citep{Wagenmaker25-dsrl}, it remains challenging to use steering for solving wholly new or heavily-randomized tasks, as we do in Bridge. Consequently, our approach cannot easily be applied to LIBERO due to deficiencies in the dataset's behavioral diversity.

\section{Training Details}
\label[appendix]{app-sec:training-details}

\begin{figure}[h]
    {\tiny\texttt{\input{text_figures/hyperparams}}}
    \caption{
        \footnotesize{Hyperparameters for training both the Steerable Policy and high-level embodied reasoner, taken from the OpenVLA parameter logging files (as both are trained by adapting the OpenVLA codebase).}
    }
    \label{fig:hyperparams}
\end{figure}

\subsection{Steerable Policy Training}
\label[appendix]{app-subsec:steerable-policy-training}

\subsubsection{OpenVLA-based Steerable Policies}
We train our first Steerable Policy by adapting the OpenVLA codebase~\citep{Kim24-openVLA}. We use all provided default hyperparameters used for training the model on the Bridge dataset (see \cref{fig:hyperparams}). Following past works~\citep{Zawalski24-ecot, Chen25-ecot-lite}, we train for 80k steps at batch size 256, split across 8x H100 GPUs on a single compute node. We use the DINOv2-SigLIP image encoder and Llama2 7b LLM version of Prismatic VLM as the base pretrained VLM for our policy~\citep{Touvron23-llama2, Oquab24-dinov2, Zhai23-siglip, Karamcheti24-prismatic}.

The main alteration to the training pipeline is that, for each frame, we identify which subtask in the overall episode it corresponds to. Then, we uniformly randomly sample a command from a list containing the overall task language provided by Bridge (to support task-level command), the subtask language, and any corresponding steering commands we generate.

Otherwise, we use the standard next-token prediction loss used for training OpenVLA. This also means the Steerable Policy expresses robot actions via naive discretization: as introduced by \citet{Brohan23-rt2}, this discretizes each of the seven action dimensions into 256 bins, which correspond to the 256 least-used tokens in Llama's vocabulary.

\subsubsection{$\pi_{0.5}$-based Steerable Policies}
Our $\pi_{0.5}$-based Steerable Policy is trained in the exact same way as above, using the default hyperparameters provided by OpenPi~\citep{Black24-pi0} for fine-tuning $\pi_{0.5}$ on the DROID dataset~\citep{Khazatsky24-droid}, albeit only run for 30k gradient steps on 8x H200. Training takes less than a day. Note that we use the standard OpenPi recipe for tuning $\pi_{0.5}$; we do NOT use the Knowledge Insulation trick used for training base $\pi_{0.5}$~\citep{Driess25-kiVLA}. Additionally, we do not condition on any proprioception information, so as to stay in line with the other policies we evaluate. Finally, we train with action chunk size 4 with end effector actions, but only execute the first action of each chunk at inference time (i.e., inference is fully closed-loop).

\subsection{High-level Embodied Reasoner Training}
\label[appendix]{app-subsec:embodied-reasoner-training}
The high-level embodied reasoner is similarly instantiated by training with an altered version of the OpenVLA codebase, using the same hyperparameters as our policy (see \cref{fig:hyperparams}). However, rather than mapping from observations and task language to actions, we instead use standard next-token prediction loss to train the VLM to predict a rationale/reasoning followed by an appropriate steering command. That is, for each training frame, we look up the corresponding subtask and rationale. Then, we also get the list of steering commands paired with that subtask, and sample one at random. Finally, we train the policy to receive the image and task language (represented as tokens), and output the rationale and the sampled command autoregressively.

Our non-reasoning ablation is trained identically, just with the rationale supervision removed (so it learns to map observations and task language directly to steering commands, without intermediate reasoning generations).

\begin{figure*}[t]
    {\tiny\texttt{\input{text_figures/generating_subtasks.tex}}}
    \caption{
        \footnotesize{The prompt for dividing Bridge tasks into subtasks.}
    }
    \label{fig:bridge-subtask-prompt}
    \vspace{0.3cm}
    {\tiny\texttt{\input{text_figures/generating_steerable_commands}}}
    \caption{
        \footnotesize{The prompt for generating steering commands for Bridge tasks.}
    }
    \label{fig:bridge-steerable-command-prompt}
    \vspace{0.3cm}
    {\tiny\texttt{\input{text_figures/generating_rationales}}}
    \caption{
        \footnotesize{The prompt for rationalizing steering commands for Bridge tasks.}
    }
    \label{fig:bridge-rationale-prompt}
\end{figure*}

\begin{figure*}[t]
    {\tiny\texttt{\input{text_figures/gemini_all_prompt}}}
    \caption{
        \footnotesize{Our approach's Gemini prompt for in-context learning VLM experiments. Note that any text in square brackets (eg. [task description]) in the prompt above is replaced with the corresponding object before the prompt is passed to the VLM.}
    }
    \label{fig:gemini-all-prompt}
    \vspace{0.3cm}
    
\end{figure*}

\begin{figure*}[t]
    {\tiny\texttt{\input{text_figures/gemini_asap_prompt}}}
    \caption{
        \footnotesize{Non-reasoning ablation's Gemini prompt for in-context learning VLM experiments. The only change from the full prompt (\cref{fig:gemini-all-prompt}) is that the VLM is instructed to answer without producing any thinking tokens. In particular, the VLM is still may output commands leveraging all levels of abstraction, and leverage in-context examples and history of past observations and commands. Note that any text in square brackets (eg. [task description]) in the prompt above is replaced with the corresponding object before the prompt is passed to the VLM.}
    }
    \label{fig:gemini-asap-prompt}
    \vspace{0.3cm}
    
\end{figure*}

\begin{figure*}[t]
    {\tiny\texttt{\input{text_figures/gemini_subtask_prompt}}}
    \caption{
        \footnotesize{SayCan-like baseline's Gemini prompt for in-context learning VLM experiments. The only change from the full prompt (\cref{fig:gemini-all-prompt}) is that the VLM is instructed to only provide subtask-level instructions. In particular, the VLM is still encouraged to reason in-context from a history of past observations and commands. Note that any text in square brackets (eg. [task description]) in the prompt above is replaced with the corresponding object before the prompt is passed to the VLM.}
    }
    \label{fig:gemini-subtask-prompt}
\end{figure*}

\begin{figure*}[t]
    {\texttt{\input{text_figures/gemini_pointing_prompt}}}
    \caption{
        \footnotesize{Prompt for extracting grounded keypoints from image observations based on a description of a target object. If a VLM instruction contains multiple keypoint descriptions, this prompt is called multiple times. Note that any text in square brackets (eg. [image observation]) in the prompt above is replaced with the corresponding object before the prompt is passed to the VLM.}
    }
    \label{fig:gemini-pointing-prompt}
\end{figure*}

\end{document}